\documentclass[10pt,twocolumn,letterpaper]{article}

\usepackage[pagenumbers]{cvpr} 

\usepackage{graphicx}
\usepackage{amsmath}
\usepackage{amssymb}
\usepackage{multirow}
\usepackage{bbm}
\usepackage{booktabs}
\usepackage{arydshln}
\usepackage{subfiles}
\usepackage{boldline}
\usepackage{term_to_use}
\usepackage{enumitem}
\usepackage{etoc}

\usepackage{xcolor}
\usepackage{color, colortbl}

\newlength\savewidth\newcommand\shline{\noalign{\global\savewidth\arrayrulewidth
  \global\arrayrulewidth 1pt}\hline\noalign{\global\arrayrulewidth\savewidth}}

\DeclareMathOperator*{\argmax}{arg\,max}
\DeclareMathOperator*{\argmin}{arg\,min}

\definecolor{defaultcolor}{gray}{0.9}
\definecolor{brightmaroon}{rgb}{0.76, 0.13, 0.28}

\newcolumntype{S}{@{}>{\lrbox0}l<{\endlrbox}}  %
\definecolor{lightgreen}{HTML}{D8ECD1}
\definecolor{lightblue}{HTML}{AFEEEE}
\newcommand{\better}[1]{\colorbox{lightblue}{#1}}
\newcommand{\cbetter}[1]{\colorbox{lightblue}{#1}}
\newcommand{\datatag}[1]{\rotatebox[origin=l]{90}{\scriptsize{#1}}}
\newcommand{\odvmm}{2}
\newcommand{\ometa}{1}
\usepackage[pagebackref,breaklinks,colorlinks]{hyperref}

\usepackage[capitalize]{cleveref}
\crefname{section}{Sec.}{Secs.}
\Crefname{section}{Section}{Sections}
\Crefname{table}{Table}{Tables}

\def\cvprPaperID{535} 
\def\confName{CVPR}
\def\confYear{2024}

\begin{document}

%
\title{MoDE: CLIP Data Experts via Clustering}

\author{Jiawei Ma$^{\ometa{}*,\odvmm{}}$ \quad Po-Yao Huang$^{\ometa{}}$ \quad Saining Xie$^{3}$ \quad Shang-Wen Li$^{\ometa{}}$ \\
Luke Zettlemoyer$^{\ometa{},4}$ \quad  Shih-Fu Chang$^{\odvmm{}}$ \quad Wen-Tau Yih$^{\ometa{}}$ \quad Hu Xu$^{\ometa{}+}$ \\
$^{1}$FAIR, Meta \quad $^{2}$Columbia University \quad $^{3}$New York University \quad $^{4}$University of Washington\\
}
\maketitle
\let\thefootnote\relax\footnotetext{$^*$ Research done while Jiawei Ma was an intern at FAIR.}
\let\thefootnote\relax\footnotetext{$^+$ Project Lead.}

\begin{abstract}
The success of contrastive language-image pretraining (CLIP) relies on the supervision from the pairing between images and captions, which tends to be noisy in web-crawled data. We present Mixture of Data Experts (MoDE) and learn a system of CLIP data experts via clustering. Each data expert is trained on one data cluster, being less sensitive to false negative noises in other clusters. At inference time, we ensemble their outputs by applying weights determined through the correlation between task metadata and cluster conditions. 
To estimate the correlation precisely, the samples in one cluster should be semantically similar, but the number of data experts should still be reasonable for training and inference. As such, we consider the ontology in human language and propose to use fine-grained cluster centers to represent each data expert at a coarse-grained level. 
Experimental studies show that four CLIP data experts on ViT-B/16 outperform the ViT-L/14 by OpenAI CLIP and OpenCLIP on zero-shot image classification but with less ($<$35\%) training cost. Meanwhile, MoDE can train all data expert asynchronously and can flexibly include new data experts. The code is available at \url{https://github.com/facebookresearch/MetaCLIP/tree/main/mode}.
\end{abstract}
\vspace{-10pt}

\section{Introduction}

Contrastive Language-Image Pretraining (CLIP) learns versatile vision-language representations which are transferable across diverse downstream tasks. Existing models, such as OpenAI CLIP~\cite{radford2021learning}, OpenCLIP~\cite{schuhmann2021laion} and MetaCLIP~\cite{xu2023demystifying}, are trained with a large collection of web-crawled image-caption pairs.
Specifically, for each image, its paired caption is viewed as a \emph{positive} example, and the captions of all the other images are viewed as \emph{negative}s.
The model then learns to project both images and captions into a shared space, where the embedding of the positive caption is drawn closer to the image embedding, compared to the embeddings of all the other negative captions.

\begin{figure}
    \centering
    \includegraphics[width=0.9\linewidth]{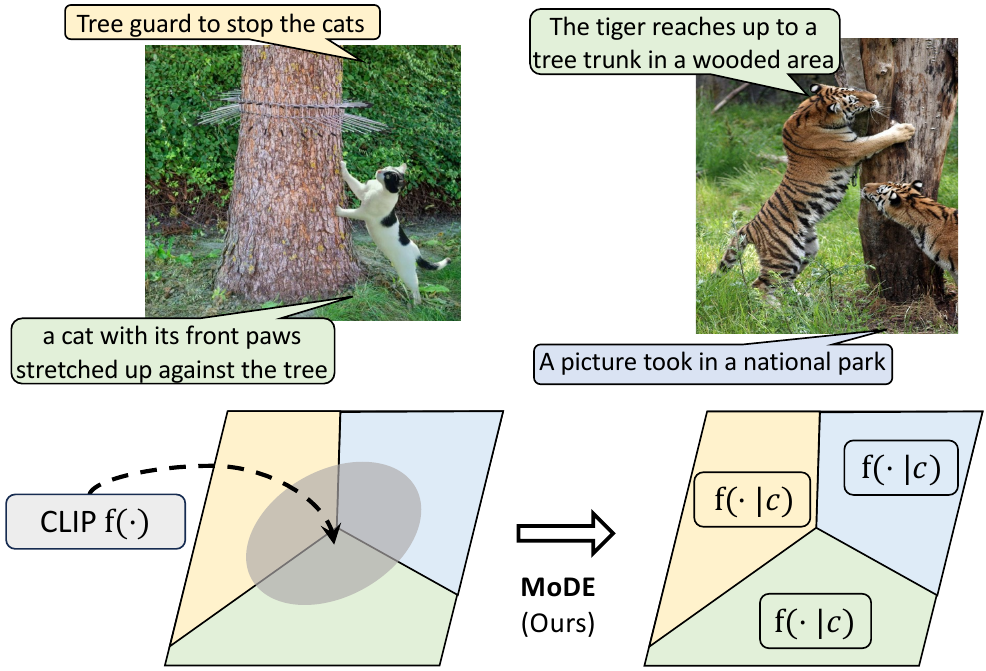}
    \caption{For an image-caption pair, the caption may describe limited visual content or even be unrelated, and such noises unavoidably hurt the quality of negative examples to learning a single model. 
    We propose to uncover the clusters from training data, where 1) the pairs with similar images but different captions are assigned to different clusters and 2) the samples in each cluster are of related meanings, and learn a \textit{Data Expert} for each cluster.
    These experts are then selectively ensembled for inference.
    }
    \label{fig:concept}
    \vspace{-15pt}
\end{figure}

The key to the success of contrastive vision-language representation learning lies in the creation of quality \emph{negative} examples for training~\cite{vsepp,pmlr-v9-gutmann10a}.
A single image can be depicted by texts with different meanings (\ie, semantics), covering multiple details and interpretations, as illustrated in \cref{fig:concept}. 
Because the paired caption usually describes limited visual content, it is common to see that two similar images have drastically different textual descriptions, especially in noisy web-crawled data. 
When those image-caption pairs are sampled in the same batch, captions of other images become \emph{false negatives} --- acceptable captions yet being treated as negative descriptions of the target image.
Conversely, if only dissimilar image-caption pairs are sampled, the contrastive learning problem becomes trivial. Incorporating \emph{hard negatives}~\cite{vsepp,videoclip,supportset} (e.g., incorrect yet similar captions that share many words of a correct textual description) in training batches has often been shown to improve the model performance.

In this work, we introduce the Mixture of Data Experts (\approach) framework (shown in Fig.~\ref{fig:concept}-bottom) via clustering. 
\approach separates false negative samples into different clusters and groups the pairs with similar semantics, which mitigates noise from false-negative captions while incorporating a more challenging set of hard-negative examples, thereby enhancing vision-language pre-training.
\approach{} consists of two main steps: (1) the training data (\ie, image-caption pairs) is first clustered into several disjoint subsets by the captions; each cluster is then used to train a model following the standard contrastive learning method. In this way, each model is specialized by the training data in one cluster and thus termed as a \textit{Data Expert}.
(2) When applied to downstream tasks, such as image classification, the task metadata (\ie, class names), are first compared to the centroid of each data cluster to determine which data expert needs to be activated. Selected data experts are then used to create the embeddings of the test image and classes. The class with the highest ensembled similarity is then output as the classification result.

Empirically,
\approach{} outperforms several state-of-the-art vision-language models when applied to multiple standard benchmarks, including +3.7\% on image classification in CLIP benchmark~\cite{radford2021learning,mu2022slip}, +3.3\% on image-to-text retrieval and +2.7\% on text-to-image retrieval on COCO~\cite{lin2014microsoft}.
The superiority of \approach can be attributed to better trained individual data expert models, due to the fact that examples in the same cluster, when used for contrastive learning, provide more quality negatives. 
Because captions in the same cluster are different but semantically similar (\eg, ``a cat climbs a tree'', ``a tiger reaches up to a tree''), they become challenging negative examples when compared with images that are not the originally paired ones.
On the other hand, it is also less likely to encounter a false negative case where a very different caption validly describes the same image (\eg, ``tree guards to stop the cats'' in \cref{fig:concept}).
\approach is also uniquely positioned for large-scale training when billions of image-caption pairs are available. As each data expert uses only a fraction of the whole dataset, it can be more easily trained with fewer compute resources asynchronously.
From experiments across different ViT~\cite{dosovitskiy2020image} model scales, we show that four ViT-B/16 data experts can outperform the single ViT-L/14 model by OpenAI CLIP~\cite{radford2021learning} and OpenCLIP~\cite{schuhmann2022laion} on image classification but requires much less ($<$35\%) training cost.
In summary, our contributions are:
\begin{itemize}[leftmargin=*]
    \item We investigate the quality \textit{negative} samples in contrastive language-image pretraining, and in particular, the noise of \textit{false negatives} in web-crawled image-caption pairs.
    \item We propose the \approach{} framework to learn a system of CLIP data experts via clustering, and adaptively ensemble data experts for downstream tasks at inference time. 
    \item Extensive experimental study has demonstrated the effects in zero-shot transfer benchmarks with low training cost. \approach{} can include new data experts flexibly and is thus beneficial for continual pre-training.
\end{itemize}

\section{Related Work}

\paragraph{Contrastive Language Image Pretraining (CLIP)}aims to learns robust \& transferable visual representations from large-scale data.
Scaling up~\cite{jia2021scaling,pham2023combined} existing approaches and improving the effectiveness is critical.
Recent progress in the field involves the exploration of regularization techniques~\cite{yu2022coca} and hyperbolic embedding methods~\cite{desai2023hyperbolic} but they require significant effort for data annotation.
Data curation is then proposed to remove noisy web-crawled image-caption pairs.
Additionally, methods like image masking~\cite{li2023scaling} and concise captions~\cite{clipa} efficiently decrease memory demands, enabling the use of larger batch sizes and model sizes.
However, a trade-off between training cost and effectiveness still exists.
Following the studies~\cite{sachidananda2023global,kalantidis2020hard} in contrastive learning~\cite{he2020momentum,chen2020simple}, recent work investigated negative samples in CLIP training but still focuses on image side~\cite{xie2023ra,liu2023learning}. The noise exhibited in captions~\cite{yang2023tempclr} is then overlooked.
In this study, we tackle the data noise and the discovery of negative samples via clustering. Rather than training a single model, 
we asynchronously train multiple data experts and then directly ensemble them for inference adaptively, which also shows benefits for model scaling.

\paragraph{Mixture-of-Expert (MoE)} trains a set of sub-models and a routing module.
Originally, each expert is defined as an entire network~\cite{jacobs1991adaptive,jordan1994hierarchical}, and a single model is selected for each data adaptively.
As restricting to hard model selection may limit the practicality, deep mixture of expert ~\cite{eigen2013learning}, is then proposed where the MoE layer is set to softly ensemble layer outputs via weighted sum,
which is then investigated with different architectures~\cite{lepikhin2020gshard,fedus2022switch} in various tasks~\cite{riquelme2021scaling,shazeer2016outrageously}.
However, all expert models are still trained on the same data simultaneously, resulting in much heavier training costs.
Recently, BTM~\cite{li2022branch,gururangan2023scaling} proposes to learn expert models on different document types (\eg, papers, posts) separately but is only validated on language models. Meanwhile, both MoE and BTM can only determine the model routing for each input separately. Instead, \approach{} generalizes to task-level adaptation and ensembles the models by task metadata (\eg, class names in classification task~\cite{deng2009imagenet}).

\begin{figure*}
    \centering
    \includegraphics[width=0.95\linewidth]{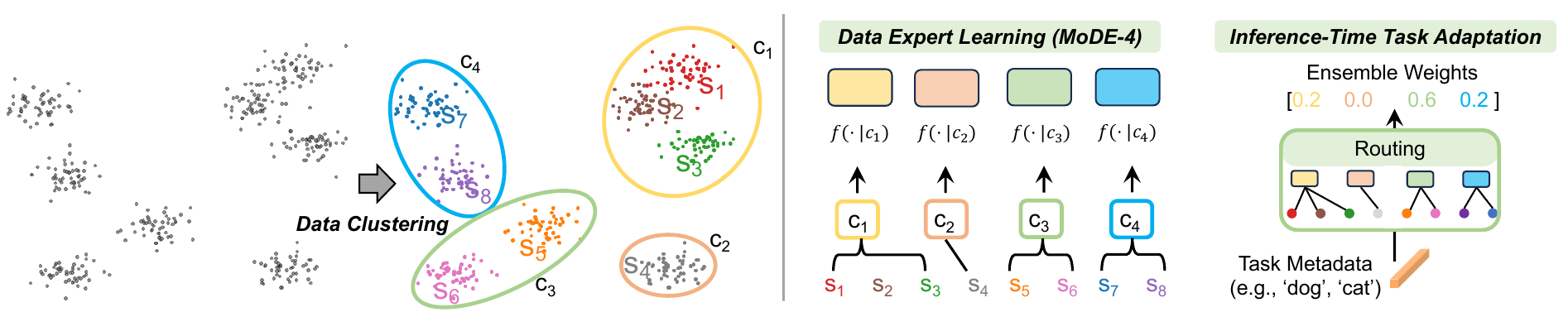} 
    \caption{Framework of \approach{} via clustering. (Left) We perform a two-step clustering on captions to decide clusters / conditions for data experts. The colored scatter plots are fine-grained clusters and the circles are clusters at coarse-grained level. (Right) Each coarse-grained cluster ($c$) conditions the learning of one data expert $f(\cdot | c)$ and all data experts (colored boxes) are learned asynchronously. For inference, the similarity between task metadata and fine-grained cluster centers ($\{s\}$) is used to decide the routing of data experts. To keep reasonable training cost, all data experts can be initialized with a model partially trained on all data without clustering (omitted for simplicity).}
    \label{fig:approach}
    \vspace{-0.5em}
\end{figure*}

\paragraph{Inference-Time Adaptation} adapts a pre-trained model quickly and effectively to new tasks. Initially, transductive learning~\cite{gammerman2013learning} is studied and leverages all unlabeled test data for model update. To mitigate the dependence on the presumed distribution of test data, test-time training\cite{sun2020test,sain1996nature,gandelsman2022test} is developed to generate individual models for each input. 
Subsequent explorations into meta-learning~\cite{snell2017prototypical,han2022few,ma2021partner} introduced a separate module (\ie, meta-learner) that can adapt the pre-trained model for each task with a few annotated examples. \approach{} has inference-time task adaptation but without annotation or parameter update.

\section{CLIP Data Experts}
For contrastive image-language pre-training, the model is trained to accurately align each image with the captions describing the visual content.
In a manner of divide-and-conquer~\cite{blahut2010fast}, for each CLIP data expert training on one cluster, we reduce the amount of false negatives and increase the hard negatives within each mini-batch. In this way, we mitigate noise exhibited in web-crawled image-caption pairs and make the model training more effective.

As shown in \cref{fig:approach}, on top of the established CLIP training that learns a single dense CLIP model $f(\cdot)$ (\cref{sec:clip}), 
we propose to learn a set of CLIP data experts $\{f(\cdot|c)\}$ via unsupervised clustering (\cref{sec:clustering}) and each CLIP data expert $f(\cdot|c)$ is trained on the cluster $c$ (\cref{sec:dataexpert}). In this way, the conditioned data expert $f(\cdot|c)$ is less sensitive to the noise from other clusters and can be effectively trained among the data of coherent semantics. 
For each evaluation task, by measuring the correlation between the task metadata (\eg, class names) and the conditions, the outputs can be jointly decided by multiple data experts (\cref{sec:inference}).

\subsection{Background: Vanilla CLIP Training}\label{sec:clip}

CLIP~\cite{radford2021learning} learns separate vision and language encoders with a joint vision-language embedding space. By contrasting positive pairs from negative samples within the same batch, CLIP can accurately model the similarity of the image and caption in each pair. We denote CLIP as $f\big((\mathbf{x}_v,\mathbf{x}_l)\big)$ for an image-caption input $(\mathbf{x}_v, \mathbf{x}_l)$, and simplify CLIP model as $f(\cdot)$.
As a reminder, instead of learning a single dense CLIP model $f(\cdot)$, we propose to learn a set of CLIP data expert models independently given a set of conditions $C$, \ie, $\{f(\cdot|c)|c\in C\}$.

\subsection{Clustering}\label{sec:clustering}

This subsection discusses how to formulate conditions $C$, and how to use clustering to automatically discover conditions for data experts from the pre-train set. In a nutshell, the desiderata for the conditions are twofold:
1) as each task at test time requires detailed description (\eg, recognize the ``cat'' species instead of just ``animal''), the conditions should be \emph{representative} such that the correlation with tasks can be precisely modeled for reliable data experts selection;
2) the number of conditions should be \emph{reasonable} since each condition is used to learn one data expert.
As each condition is represented by a cluster, the ideals of \emph{representative} likely ask for more fine-grained clustering whereas the latter may require for fewer data experts.

Instead, motivated by the ontology in human language, we propose to capture such a hierarchical structure via clustering, \ie, determine the condition of a data expert at the coarse-grained level and represent it via the set of fine-grained clusters.
For simplicity, we design a \textit{two-step K-means clustering}. 
We first employ fine-grained clustering to locate each cluster whose samples are of similar semantics, such that the fine-grained cluster centers are representative (Step 1),
and then group fine-grained clusters to determine coarse-grained clustering among data for data experts' specialization (Step 2). 
In this way, instead of using a single coarse-grained center, the condition is symbolized by the fine-grained cluster centers. 
The features for clustering are extracted from captions and the details are studied in \cref{sec:discussion}.

\paragraph{Step 1: Fine-grained Clustering.}
As the amount of pre-train data $\mathcal{D}$ is huge (hundreds of millions to billions level for CLIP~\cite{radford2021learning}), it could be inefficient to train K-means over all pre-training data. Instead, we first uniformly sample a subset from the pre-training set: $\mathcal{D}' \sim \mathcal{D}$ and $|\mathcal{D}'| \ll |\mathcal{D}|$.
Then, we perform K-means training ~\cite{mitchell1997machine} over $\mathcal{D}'$: 
\begin{equation}
S \leftarrow \text{K-means}(\mathcal{D}'),  
\end{equation}
where $S$ is a set of learned cluster centers.
Note that the number of fine-grained clusters $m=|S|$ can be substantially large such that the cluster center of each cluster well represents coherent semantic information for each cluster. 

\noindent \textbf{Step 2: Coarse-grained Clustering.}To efficiently allocate the training/inference of a data expert, we perform a second round, \ie, coarse-grained, K-means clustering on top of fine-grained cluster centers $S$: 
\begin{equation}C\leftarrow\text{K-means}(S),
\end{equation}
where each coarse-grained cluster center $c\in C$ is the condition for a data expert.
We denote $n=|C|$ as the number of data experts where $n \ll m$, and $S_c$ as set of fine-grained clusters assigned to the data expert $f(\cdot|c)$ where $S = \cup_{c \in C} S_c$.

\subsection{Data Experts Training}\label{sec:dataexpert}

Next, we formulate training data for each data expert.
We first 
collect the data assigned for each fine-grained cluster $s$: 
$\mathcal{D}_s = \{d|s=\argmin_{s\in S}(\|\mathbf{e}_d - \mathbf{e}_s\|_2^2)\text{ and } d \in \mathcal{D} \}$, 
where $\mathbf{e}_d$ and $\mathbf{e}_s$ are the embeddings for training example $d$ and fine-grained cluster center $s$ respectively.
To train a data expert $f(\cdot|c)$, its corresponding CLIP training data is: 
\begin{equation}
\mathcal{D}_c = \bigcup\nolimits_{s \in S_c} \mathcal{D}_s.    
\end{equation}
For convenience, we use \approach{-n} to indicate the system with $n$ CLIP data experts.
For training efficiency, all data experts are specialized from the same seed CLIP model that is partially trained  over the entire set $\mathcal{D}$. Then, each data expert $f( \cdot | c)$ is trained only on $\mathcal{D}_c$.

\begin{table*}[!htb]  
\centering
\vspace{-1.em}
\setlength\tabcolsep{1.0pt}
\resizebox{\linewidth}{!}{
\begin{tabular}{lc|cccccccccccccccccccccccccc}
& \datatag{Average}
& \datatag{ImageNet}
& \datatag{Food-101} 
& \datatag{CIFAR10} 
& \datatag{CIFAR100} 
& \datatag{CUB}
& \datatag{SUN397}
& \datatag{Cars}
& \datatag{Aircraft}
& \datatag{DTD}
& \datatag{Pets}
& \datatag{Caltech-101}
& \datatag{Flowers}
& \datatag{MNIST}
& \datatag{FER-2013}
& \datatag{STL-10}
& \datatag{EuroSAT}
& \datatag{RESISC45}
& \datatag{GTSRB}
& \datatag{KITTI}
& \datatag{Country211}
& \datatag{PCAM}
& \datatag{UCF101}
& \datatag{Kinetics700}
& \datatag{CLEVR}
& \datatag{HatefulMemes}
& \datatag{SST2}\\
\shline
\rowcolor{lightgray} \multicolumn{28}{l}{ViT-B/32}\\
OpenAI CLIP     & 56.6 & 63.4 & \better{83.7} & 89.8 & 65.1 & 53.7 & 62.0 & 59.7 & 19.6 & 44.0 & 87.2 & 87.4 & 66.9 & \better{48.2} & 46.6 & \better{97.1} & 44.9 & 61.0 & 32.6 & 28.7 & 17.2 & \better{62.5} & 63.9 & \better{48.0} & 23.6 & 56.4 & \better{58.6} \\
OpenCLIP & 57.6 & 62.9 & 80.7 & 90.7 & 70.6 & 61.2 & \better{66.4} & \better{79.2} & 16.7 & \better{54.5} & 86.5 & 90.7 & 66.1 & 37.4 & \better{48.2} & 95.6 & 52.2 & 58.0 & \better{42.0} & \better{38.0} & 14.8 & 50.1 & 63.0 & 42.8 & 22.5 & 53.3 & 52.3 \\
\baselinetable{} & 58.2 & 65.5 & 80.6 & \better{91.3} & 70.2 & 63.4 & 63.0 & 70.7 & 26.8 & 52.8 & 88.7 & 91.9 & 68.5 & 41.5 & 35.9 & 95.4 & 52.6 & \better{64.2} & 35.8 & 30.7 & 17.2 & 55.5 & 66.1 & 45.4 & \better{30.6} & 56.4 & 53.4 \\ \hline
\approach{}-2 & 58.6 & 66.1 & 81.2 & 90.9 & 70.5 & 65.2 & 63.0 & 72.0 & 28.3 & 53.5 & 89.4 & \cbetter{92.3} & 68.2 & 45.2 & 33.5 & 95.4 & 51.9 & 63.7 & 34.9 & 34.2 & \cbetter{17.3} & 54.3 & 65.9 & 45.5 & 29.3 & 56.6 & 54.6 \\
\approach{}-4 & \better{59.0} & \cbetter{66.4} & 82.3 & \cbetter{91.3} & \cbetter{70.9} & \cbetter{67.0} & 63.7 & 73.8 & \cbetter{30.1} & 52.6 & \cbetter{89.9} & 92.1 & \cbetter{69.2} & 37.9 & 33.2 & 95.7 & \cbetter{53.5} & {64.1} & 35.2 & 33.9 & 17.1 & 58.4 & \cbetter{66.6} & 45.9 & 30.0 & \cbetter{58.0} & 54.5 \\ \shline
\rowcolor{lightgray} \multicolumn{28}{l}{ViT-B/16}\\
OpenAI CLIP     & 59.6 & 68.3 & \better{88.8} & 90.8 & 68.2 & 55.6 & 64.0 & 64.6 & 24.0 & 45.1 & 88.9 & 89.1 & 69.4 & 51.8 & \better{53.0} & \better{98.2} & 54.8 & 65.5 & 43.3 & 21.7 & 22.8 & \better{56.3} & 68.5 & \better{52.3} & 25.5 & \better{58.7} & 60.5 \\
OpenCLIP & 60.4 & 67.1 & 85.8 & \better{91.7} & 71.4 & 65.3 & \better{69.2} & \better{83.6} & 17.4 & 51.0 & 89.2 & 90.8 & 66.5 & \better{66.3} & 46.1 & 97.0 & 52.2 & 65.7 & 43.5 & 23.7 & 18.1 & 51.7 & 67.0 & 46.2 & \better{33.9} & 54.5 & 54.4 \\
\baselinetable{} & 61.1 & 70.8 & 86.8 & 90.1 & 66.5 & 70.8 & 66.6 & 74.1 & 27.9 & 55.9 & 90.4 & 93.8 & 72.3 & 47.8 & 44.6 & 97.2 & \better{55.4} & 68.8 & 43.8 & \better{33.4} & 22.6 & 52.9 & 68.0 & 49.5 & 22.8 & 54.8 & \better{60.6} \\ \hline
\approach{}-2 & 61.8 & 71.2 & 87.2 & 91.3 & 67.4 & 71.7 & 66.8 & 75.5 & 29.9 & \cbetter{57.0} & 90.5 & \cbetter{94.1} & 73.0 & 51.0 & 44.9 & 97.2 & \cbetter{55.4} & 68.7 & \cbetter{44.5} & 32.9 & 22.7 & 52.9 & 67.2 & 49.4 & 28.1 & 56.0 & 60.1 \\
\approach{}-4 & \better{62.1} & \cbetter{71.6} & 87.8 & 91.4 & \better{68.9} & \cbetter{74.7} & 67.2 & 77.3 & \cbetter{32.6} & 56.2 & \cbetter{91.3} & 93.9 & \cbetter{74.9} & 43.7 & 46.6 & 97.2 & 54.4 & \cbetter{70.0} & 44.0 & 29.8 & \better{22.9} & 55.7 & \better{68.6} & 50.0 & 29.7 & 55.2 & 58.0 \\ \shline
\rowcolor{lightgray} \multicolumn{28}{l}{ViT-L/14}\\
OpenAI CLIP     & 65.7 & 75.5 & \better{93.0} & 95.6 & \better{78.3} & 63.3 & 66.8 & 77.8 & 31.3 & 55.3 & 93.6 & 93.3 & 79.3 & \better{76.4} & \better{56.9} & \better{99.4} & 61.9 & 70.9 & \better{50.6} & 19.2 &\better{31.9} & 50.1 & 75.7 & \better{60.2} & 22.3 & \better{59.7} & 68.9 \\
OpenCLIP & 64.5 & 72.7 & 90.0 & 94.7 & 78.0 & 73.9 & \better{72.4} & \better{89.5} & 24.7 & 60.2 & 91.6 & 93.6 & 73.0 & 76.1 & 54.3 & 98.1 & \better{63.9} & 69.6 & 49.9 & 16.0 & 23.0 & 51.7 & 71.5 & 51.6 & 25.4 & 55.3 & 56.0 \\
\baselinetable{} & 67.1 & 76.2 & 90.7 & 95.5 & 77.4 & 75.9 & 70.5 & 84.7 & 40.4 & 62.0 & 93.7 & 94.4 & 76.4 & 61.7 & 46.5 & 99.3 & 59.7 & 71.9 & 47.5 & \better{29.9} & 30.9 & 70.1 & 75.5 & 57.1 & \better{35.1} & 56.6 & \better{65.6} \\ \hline
\approach{}-2 & 67.1 & \better{76.5} & 91.1 & \better{95.9} & 77.8 & 76.7 & 70.6 & 85.1 & 40.9 & \better{62.4} & 93.9 & \better{94.8} & 76.8 & 63.0 & 46.2 & \better{99.4} & 57.8 & 71.7 & 47.4 & 26.7 & 31.1 & 69.9 & 75.6 & 57.3 & 33.1 & 56.6 & 65.5 \\
\approach{}-4 &   \better{67.2} & 76.3 & 91.2 & 95.7 & 77.9 & \better{78.3} & 70.7 & 85.6 & \better{41.8} & \better{62.4} & \better{94.0} & 94.5 & \better{77.1} & 62.6 & 46.6 & 99.2 & 57.7 & \better{72.0} & 47.3 & 26.8 & 31.3 & \better{71.5} & \better{76.0} & 57.3 & 30.6 & 56.6 & 65.5 \\
\shline
\end{tabular}
}
\caption{Performance on CLIP benchmark~\cite{radford2021learning,mu2022slip} by models trained on 400M image-caption pairs. \approach{-2} and \approach{-4} consistently outperform the \baseline{} and \approach{-4} achieves the best score on average.}
\label{tab:clip400m}
\end{table*}

\subsection{Inference Time Task-Adaptation}\label{sec:inference}

As our framework conditions the model expertise on clusters to train data experts,
it also gives multiple models to choose from during inference (instead of the only choice on a single CLIP model). This gives the room to adapt different data experts to various downstream tasks.

We propose a simple approach to adapt data experts (no parameter updates) to downstream tasks using the task metadata.
Intuitively, 
this approach routes each downstream task adaptively and efficiently to data experts during inference.  
For simplicity, we formulate the data experts routing as a weighted sum of data experts' outputs. Formally, given an evaluation task $\mathbf{T}$, the output of CLIP data experts is
\begin{equation}
    \sum\nolimits_{c\in C} f(\cdot | c) p(c|\mathbf{T}) ,
\end{equation}
where $p(c|\mathbf{T})$ is the normalized weight for the data expert $f( \cdot | c)$, \ie, $\sum\nolimits_{c\in C} p(c|\mathbf{T}) = 1$.
The weight is proportional to the correlation, \ie, similarity, between metadata of task $\mathbf{T}$ and condition $c$.
Below we provide simple implementations for zero-shot classification and retrieval, respectively.

\bitem{Zero-Shot Classification.} 
To have accurate routing, we leverage fine-grained cluster centers $S$ in Step 1 to route a task to data experts.
We treat the set of class names $L$ as metadata, and define the similarity matrix between classes and data experts as 
$\mathbf{A} \in \mathbb{R}^{|L| \times m}$.
To compute $\mathbf{A}$, we first compute $\mathbf{e}_l$ as the embedding for class $l\in L$ via the same encoder for the embedding of fine-grained cluster center $\mathbf{e}_s$.
Then each entry is defined as 
\begin{equation}\label{eq:closeness}
    \mathbf{A}_{l,s} = \exp(-\|\mathbf{e}_l - \mathbf{e}_{s}\|_2^2 / \lambda),
\end{equation}
where $\lambda{} \in \mathbb{R}^+$ is a temperature to sharpen the similarities. Further, the weight routing to a data expert $f( \cdot | c)$ is proportional to 
\begin{equation}\label{eq:aggregation} 
    p(c|\mathbf{T}) \propto \exp(\sum\nolimits_{l\in L}\sum\nolimits_{s \in S_c}\mathbf{A}_{l,s}).
\end{equation}
In practice, we found that using the nearest neighboring fine-grained cluster center ($\argmax_{s\in S} \mathbf{A}_{l, s}$) for each class $l \in L$ is good enough to reduce noises in routing.

\bitem{Zero-Shot Retrieval}. The retrieval tasks consist of text retrieval and image retrieval. For text retrieval where each image is used to retrieve a text from a large corpus $Q$, we leverage $Q$ as metadata to build similarity matrix $\mathbf{A}\in \mathbb{R}^{|Q| \times m}$. 
Similar to the classification task, the weights for ensembling can be naturally adopted for \approach{}:
\begin{equation}
    p(c|\mathbf{T}) \propto \exp(\sum\nolimits_{q\in Q}\sum\nolimits_{s \in S_c}\mathbf{A}_{q,s}),
\end{equation}
where each entry $\mathbf{A}_{q,s}$ is computed as $\exp(-\|\mathbf{e}_q - \mathbf{e}_{s}\|_2^2 / \lambda)$, where $\mathbf{e}_q$ is the embedding for text $q$.
For image retrieval where each text $q$ retrieves an image separately, we treat the retrieval by text $q$ as an independent task $\mathbf{T}_q$ such that the ensembling weights are then $p(c|\mathbf{T}_q) \propto \exp(\sum\nolimits_{s \in S_c}\mathbf{A}_{q,s})$.

\begin{table*}[!htb]
\vspace{-1.em}
\centering
\setlength\tabcolsep{1.0pt}
\resizebox{\linewidth}{!}{
\begin{tabular}{lc|cccccccccccccccccccccccccc}
& \datatag{Average}
& \datatag{ImageNet}
& \datatag{Food-101} 
& \datatag{CIFAR10} 
& \datatag{CIFAR100} 
& \datatag{CUB}
& \datatag{SUN397}
& \datatag{Cars}
& \datatag{Aircraft}
& \datatag{DTD}
& \datatag{Pets}
& \datatag{Caltech-101}
& \datatag{Flowers}
& \datatag{MNIST}
& \datatag{FER-2013}
& \datatag{STL-10}
& \datatag{EuroSAT}
& \datatag{RESISC45}
& \datatag{GTSRB}
& \datatag{KITTI}
& \datatag{Country211}
& \datatag{PCAM}
& \datatag{UCF101}
& \datatag{Kinetics700}
& \datatag{CLEVR}
& \datatag{HatefulMemes}
& \datatag{SST2}\\
\shline
\rowcolor{lightgray} \multicolumn{28}{l}{ViT-B/32}\\
OpenCLIP & 61.5 & 66.6 & 82.0 & 93.6 & 75.8 & 66.0 & \better{68.3} & \better{86.0} & 23.9 & 56.1 & 90.5 & 91.9 & 70.5 & \better{70.0} & 50.4 & 96.6 & 49.3 & 65.7 & 49.3 & \better{32.7} & 16.7 & \better{51.7} & 64.9 & 45.6 & \better{24.2} & 52.4 & \better{57.2} \\
\baselinetable{} & 59.8 & 67.6 & 82.6 & 95.2 & 77.7 & 67.8 & 66.8 & 77.2 & 26.9 & 58.9 & 90.9 & 92.5 & 69.7 & 42.7 & 48.3 & 96.3 & 49.9 & 66.5 & 39.2 & 29.3 & 17.7 & 50.0 & 68.0 & 47.6 & 19.4 & \cbetter{53.5} & 53.1 \\ \hline
\approach{-2} & 61.2 & 68.7 & 84.1 & \better{95.3} & 78.6 & 69.5 & 67.0 & 80.8 & \cbetter{30.9} & \better{60.6} & 91.0 & \better{92.9} & 71.9 & 40.8 &\better{50.4} & 96.3 & \better{51.3} & \cbetter{67.9} & 44.2 & 31.4 & 18.3 & \better{51.3} & \cbetter{69.0} & 47.4 & 23.2 & 52.6 & 54.4 \\
\approach{-4} & \better{61.7} & \better{68.8} & \better{85.8} & 95.2 & \better{79.0} & \better{74.4} & 67.5 & 83.3 & 29.5 & 60.3 & \better{91.9} & \better{92.9} & \better{72.1} & 49.7 & 46.9 & \better{96.4} & 50.3 & 66.8 & \better{51.6} & \better{28.5} & \better{19.6} & 50.1 & 68.4 & \better{48.3} & 21.6 & 52.6 & 52.2 \\  \shline
\rowcolor{lightgray} \multicolumn{28}{l}{ViT-B/16}\\
OpenCLIP & 62.4 & 70.2 & 86.2 & 94.9 & 76.9 & 70.5 & 70.6 & 88.2 & 26.6 & 56.3 & 90.4 & 93.1 & 71.0 & 65.8 & \better{53.3} & 97.9 & \better{55.2} & 68.3 & 48.3 & 11.9 & 20.3 & 51.2 & 68.1 & 48.9 & 24.8 & 53.0 & \better{59.5} \\
\baselinetable{} & 63.5 & 72.1 & 88.3 & 95.7 & 79.0 & 71.4 & 68.5 & 82.9 & 30.3 & 62.1 & 91.7 & 93.3 & 73.9 & 66.1 & 47.0 & 98.4 & 51.1 & 71.1 & 46.6 & 16.6 & 22.7 & 50.5 & 73.0 & 52.5 & 30.8 & 57.4 & 59.0 \\ \hline
\approach{-2} & 65.0 & 73.6 & 89.5 & 96.0 & 81.4 & 76.5 & 69.0 & 85.7 & 35.9 & \better{63.5} & 93.4 & 93.4 & 75.5 & 59.2 & 46.4 & 98.3 & 50.0 & \better{72.0} & 50.1 & \better{34.9} & 23.9 & 50.8 & 71.2 & 52.1 & \better{31.2} & \better{59.1} & 58.4 \\
\approach{-4} & \better{67.2} & \better{74.2} & \better{91.6} & \better{96.5} & \better{82.0} & \better{80.9} & \better{71.2} & \better{88.9} & \better{42.2} & 63.0 & \better{93.6} & \better{93.6} & \better{78.9} & \better{66.8} & 49.0 & \better{98.5} & 53.8 & 71.5 & \better{57.5} & 32.4 & \better{26.7} & \better{61.7} & \better{73.8} & \better{53.9} & 27.4 & 57.0 & 59.4 \\ \shline 
\rowcolor{lightgray} \multicolumn{28}{l}{ViT-L/14}\\
OpenCLIP & 65.7 & 74.0 & 88.6 & 95.8 & 78.3 & 73.5 & 73.5 & \better{91.4} & 34.6 & 61.2 & 92.7 & 93.3 & 74.4 & 64.4 & 53.9 & 98.5 & 58.6 & 71.9 & 51.6 & 26.1 & 24.4 & 58.0 & 73.3 & 52.0 & \better{27.4} & 55.1 & 60.4 \\
\baselinetable{} & 69.8 & 79.2 & 93.4 & 97.6 & 84.2 & 80.1 & 73.8 & 88.7 & 44.6 & 68.1 & 94.7 & 95.4 & \better{81.8} & 64.4 & \better{55.1} & \better{99.3} & 59.2 & 74.6 & 56.3 & 29.7 & \better{34.0} & 67.3 & 81.6 & 62.0 & 25.9 & \better{58.0} & \better{66.7} \\ \hline
\approach{-2} & 70.4 & \better{79.5} & 93.5 & 97.6 & 85.0 & 82.9 & 74.0 & 90.9 & \better{49.0} & \better{69.5} & 95.0 & 95.3 & \better{81.8} & 69.7 & 53.7 & 99.2 & \better{63.3} & 75.2 & 59.0 & 29.8 & 33.9 & 62.3 & \better{81.7} & \better{62.4} & 24.0 & 56.6 & 65.4\\
\approach{-4} &  \better{71.2} & 79.4 & \better{94.0} & \better{97.8} & \better{85.6} & \better{83.5} & \better{74.2} & 91.2 & 48.7 & 69.1 & \better{95.6} & \better{95.6} & 81.4 & \better{71.4} & 54.3 & \better{99.3} & 61.0 & \better{76.5} & \better{63.3} & \better{34.7} & \better{34.0} & \better{70.9} & 81.6 & 62.2 & 24.6 & 55.7 & \better{66.7} \\
\shline
\end{tabular}
}
\caption{Performance on CLIP benchmark~\cite{radford2021learning,mu2022slip} by models trained on billion-scale dataset (OpenCLIP: 2.3B, \baselinetable{}/\approach{}: 2.5B). \approach{-2} and \approach{-4} consistently outperform the \baseline{} and \approach{-4} achieves the best score on average.}
\label{tab:clip2b5}
\end{table*}

\section{Experiment}\label{sec:exp}

\subsection{Data}
We use the datasets collected in MetaCLIP~\cite{xu2023demystifying} for evaluation and conduct experiments on image-caption pairs at two scales: 400M (similar to the scale in OpenAI CLIP), and 2.5B to scale \approach{}.
All images are pre-processed with face-blurring and de-duplication against benchmarks.

\subsection{Training Setup}
\bitem{Clustering Setup.} We use the pre-trained language model SimCSE~\cite{gao2021simcse} to extract the embeddings for all captions 
where the advantages of language encoders over CLIP encoders are studied in \cref{sec:embedding_type}.
We use balanced K-means~\cite{malinen2014balanced} for both of the two unsupervised clustering steps.
We set the number of fine-grained clusters $m=1024$, and report performance for both \approach{-2} and \approach{-4} below to directly show the improvement by increase the number of data expert models on all evaluation tasks.

\bitem{Data Experts Training Setup.} We follow OpenAI CLIP's hyper-parameters~\cite{radford2021learning} for fair comparison and train on the same budget of 12.8B image-caption pairs (32 epochs of 400M), with a global batch size of 32,768. 
We train \approach{} under 3 scales: for ViT-B/32 and ViT-B/16, we use 64 Nvidia V100 GPUs with a per GPU batch size of 512, and for ViT-L/14, we use 128 GPUs with a 256 per GPU batch size.
To maintain a reasonable training cost, we start MoDE training from the 27th epoch (out of 32 epochs) of a partially trained MetaCLIP as the seed model and all data experts share the same seed model to save computes.

\subsection{Evaluation}
\bitem{Zero-Shot Image Classification.} 
We follow the evaluation protocol in CLIP benchmark~\cite{mu2022slip,radford2021learning,xu2023demystifying} and use the same class names \& prompts by OpenAI CLIP.
For fair comparison, MetaCLIP~\cite{xu2023demystifying} naturally serves as the single dense baseline.
The checkpoints of OpenAI CLIP (WIT400M data)~\cite{radford2021learning} and OpenCLIP (LAION-400M data, LAION-2B data)~\cite{schuhmann2021laion} are also re-evaluated for fair comparison.

\begin{table*}[ht]
    \centering\footnotesize
    \parbox{.49\linewidth}{
        \centering
        \footnotesize
        \setlength\tabcolsep{0.8pt}
        \resizebox{0.99\linewidth}{!}
        {\renewcommand{\arraystretch}{1.03}      
        \begin{tabular}{cc|cccccc|cccccc}
        \shline
        Approach & ViT & Avg. & IN-Sk & IN-V2 & IN-A & IN-O & IN-R & Avg. & IN-Sk & IN-V2 & IN-A & IN-O & IN-R \\ \hline
        OpenAI CLIP     & \multirow{5}{*}{B/32} & 49.4 & 42.3 & 56.0 & \better{31.5} & 47.8 & 69.4 &  -  &   -  &  -  &  -  &  -  &  -  \\
        OpenCLIP & & 50.6 & 49.4 & 55.1 & 21.7 & \better{53.5} & 73.4 & 52.9 & 53.7 & 58.1 & 26.3 & \better{50.0} & 76.4 \\
        \baselinetable{} & & 52.2 & 53.3 & 57.6 & 28.6 & 46.8 & 74.8 & 54.4 & 56.0 & 59.6 & 29.9 & 48.3 & 78.1 \\
        \approach{-2} & & 53.0 & 53.9 & 57.9 & 29.4 & 48.0 & 75.7 & 55.2 & 57.1 & 60.5 & 31.2 & 48.4 & 79.0 \\
        \approach{-4} & & \better{53.4} & \better{54.4} & \better{58.5} & 30.8 & 47.6 & \better{76.0} & \better{56.5} & \better{57.6} & \better{61.6} & \better{34.2} & 49.2 & \better{80.0} \\ \hline
        OpenAI CLIP     & \multirow{5}{*}{B/16} & 56.0 & 48.3 & 61.9 & \better{50.0} & 42.3 & 77.7 &   -  &   -  &  -  &  -  &  -  &  -  \\
        OpenCLIP & & 54.8 & 52.4 & 59.7 & 33.2 & \better{50.7} & 77.9 & 56.7 & 56.1 & 62.3 & 38.2 & \better{46.3} & 80.6 \\
        \baselinetable{} & & 57.7 & 57.9 & 62.6 & 47.0 & 39.2 & 81.8 & 60.1 & 60.2 & 65.0 & 49.5 & 41.6 & 84.2 \\
        \approach{-2} & & 58.4 & 58.5 & 63.2 & 47.9 & 39.9 & 82.3 & 62.3 & 62.4 & 66.5 & 52.0 & 45.2 & 85.5 \\
        \approach{-4} & & \better{59.0} & \better{58.8} & \better{63.7} & 49.2 & 40.4 & \better{82.9} &  \better{63.3} & \better{62.8} & \better{67.1} & \better{55.7} & 44.5 &  \better{86.6} \\  \hline
        OpenAI CLIP     & \multirow{5}{*}{L/14} & \better{64.1} & 59.6 & 69.8 & \better{70.7} & 32.3 & 87.9 &  -  &   -  &  -  &  -  &  -  &  -  \\
        OpenCLIP & & 59.6 & 59.6 & 65.5 & 46.5 & \better{42.0} & 84.7 & 62.2 & 63.3 & 67.8 & 53.9 & \better{38.7} & 87.4 \\
        \baselinetable{} & & 63.8 & 65.0 & 69.8 & 66.4 & 28.9 & 88.9 & 67.2 & 68.9 & 72.6 & 72.3 & 30.2 & 92.1 \\
        \approach{-2} & & 64.0 & 65.2 & 70.0 & 66.9 & 28.9 & \better{89.0} & 67.6 & 69.3 & 72.8 & 73.0 & 30.6 & 92.3 \\
        \approach{-4} & & \better{64.1} & \better{65.3} & \better{70.1} & 66.8 & 29.4 & \better{89.0} & \better{68.2} & \better{69.9} & \better{73.3} & \better{74.0} & 31.3 & \better{92.7} \\ \shline
        \rowcolor{lightgray} \multicolumn{2}{c|}{Pre-Train Data} & \multicolumn{6}{c|}{400M Image-Caption Pairs} & \multicolumn{6}{c}{OpenCLIP:2.3B; \baselinetable{}/MoDE:2.5B} \\ 
        \end{tabular}}
        \caption{\textbf{Zero-Shot Robustness Evaluation}. The results are separated by the scale of pre-train set. Entries in blue are the best ones.}\label{tab:robustness}
        
    }
    \hfill
    \parbox{.49\linewidth}{
        \centering
        \footnotesize
        \setlength\tabcolsep{0.8pt}
        \resizebox{0.99\linewidth}{!}
        {\renewcommand{\arraystretch}{1.1}
        \begin{tabular}{cc|cccccc|cccccc}
        \shline
        \multirow{3}{*}{Approach} & \multirow{3}{*}{ViT} & \multicolumn{6}{c|}{Text Retrieval}                       & \multicolumn{6}{c}{Image Retrieval}                      \\
                                  &                        & \multicolumn{3}{c}{COCO} & \multicolumn{3}{c|}{Flickr30k} & \multicolumn{3}{c}{COCO} & \multicolumn{3}{c}{Flickr30k} \\ 
                                  &                        & R@1      & R@5      & R@10    & R@1    & R@5    & R@10   & R@1      & R@5      & R@10    & R@1    & R@5    & R@10   \\ \hline
        OpenCLIP               & \multirow{4}{*}{B/32} & 56.3 & 79.8 & 87.1 & \better{84.1} & \better{96.2} & \better{98.3} & 39.3 & 65.4 & 75.6 & \better{66.7} & {88.4} & 93.1 \\
        \baselinetable{}          &                        & 55.2 & 78.9 & 86.5 & 80.7 & 95.2 & 97.3 & 38.1 & 64.1 & 74.3 & 65.1 & 87.7 & 92.7 \\
        \approach{-2}           &                        & 56.7 & \better{80.2} & \better{87.5} & 82.8 & 95.1 & 98.2 & 39.5 & 65.3 & 75.3 & 66.4 & \better{89.0} & \better{93.6} \\
        \approach{-4}           &                        & \better{57.4} & {80.1} & 87.3 & 82.9 & 95.6 & 97.7 & \better{39.9} & \better{66.1} & \better{75.7} & \better{66.7} & {88.4} & {93.3} \\ \hline
        OpenCLIP               & \multirow{4}{*}{B/16} & 59.5 & 81.8 & 88.6 & 86.2 & \better{98.0} & 99.5 & 42.3 & 67.7 & 77.1 & 69.8 & 90.4 & 94.6 \\
        \baselinetable{}            &                        & 59.4 & 80.6 & 87.8 & 85.5 & 97.4 & 98.9 & 41.4 & 67.2 & 76.9 & 70.7 & 90.8 & 94.5 \\
        \approach{-2}           &                        & 60.7 & 82.6 & 89.0 & 87.3 & 97.6 & 99.2 & 43.1 & 68.6 & 77.8 & 72.1 & \better{91.8} & 95.3 \\
        \approach{-4}           &                        & \better{62.7} &\better{82.9} & \better{89.8} & \better{89.4} & \better{98.0} & \better{99.6} & \better{44.1} & \better{69.5} & \better{78.7} & \better{72.6} & \better{91.8} & \better{95.4}  \\ \hline
        OpenCLIP  & \multirow{4}{*}{L/14}   & 63.3 & 83.9 & 90.8 & 89.5 & 98.7 & 99.4 & 46.5 & 71.1 & 79.8 & 75.5 & 92.9 & 95.9 \\
        \baselinetable{} &  & 64.4 & 85.0 & 91.3 & 90.1 & 98.6 & 99.3 & 47.1 & 71.4 & 80.3 & 76.5 & 93.6 & 96.5 \\
        \approach{-2} & & 65.2 & 85.3 & 91.6 & 90.9 & 98.9 & 99.6 & 47.9 & 72.1 & 80.6 & 77.2 & \better{93.7} & 96.6 \\
        \approach{-4} & & \better{65.5} & \better{85.4} & \better{91.8} & \better{91.2} & \better{99.0} & \better{99.7} & \better{48.2} & \better{72.4} & \better{80.7} & \better{77.6} & \better{93.7} & \better{96.7} \\ \shline
        \rowcolor{lightgray} \multicolumn{2}{c|}{Pretrain Data}&\multicolumn{12}{c}{OpenCLIP:2.3B; \baselinetable{}/MoDE:2.5B}\\ 
        \end{tabular}}
        \caption{\textbf{Zero-shot Retrieval}. Entries in blue are the best ones. Results by model trained on 400M pairs can be found in the Suppl.}
        \label{tab:retrieval}
    }    
\end{table*}

\begin{figure}
    \centering
    \includegraphics[width=0.9\linewidth]{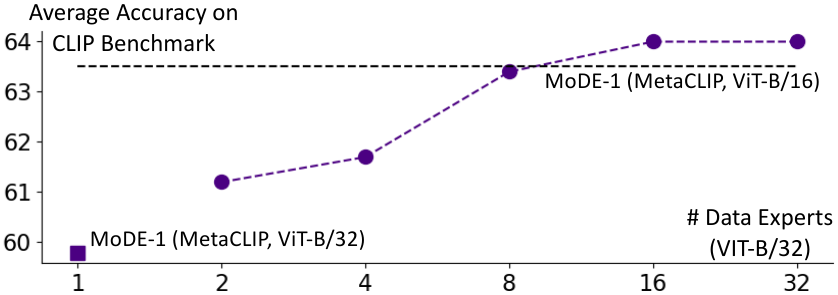} 
    \caption{Average accuracy CLIP benchmark with increased number of data expert models in \approach{} (Pretrain set: 2.5B pairs).}
    \label{fig:scaling-expert}
\end{figure}

The framework \approach{} has shown \emph{consistent performance gain across model scales and data scales}. 
Firstly, we compare the models learned from 400M-scale dataset in \cref{tab:clip400m},
and summarize the results by different model scales. \approach{} achieves consistent performance gain where increasing the number of data experts results in better performance.
Next, we study the scaling property of MoDE on 2.5B image-text pairs. From \cref{tab:clip2b5}, comparing against MetaCLIP~\cite{xu2023demystifying}, the advantage of \approach{} to learn four data expert models is better revealed on scaling training data: +1.9\% on B/32,  +3.7\% on B/16, and +1.4\% on L/14. 
Lastly, we increase the number of data experts. As shown in \cref{fig:scaling-expert}, the performance can be kept improving when we increase the number of data experts where \approach{-16} ViT-B/32 can outperform the MetaCLIP ViT-B/16 baseline.

\begin{figure}
    \centering
    \includegraphics[width=0.9\linewidth]{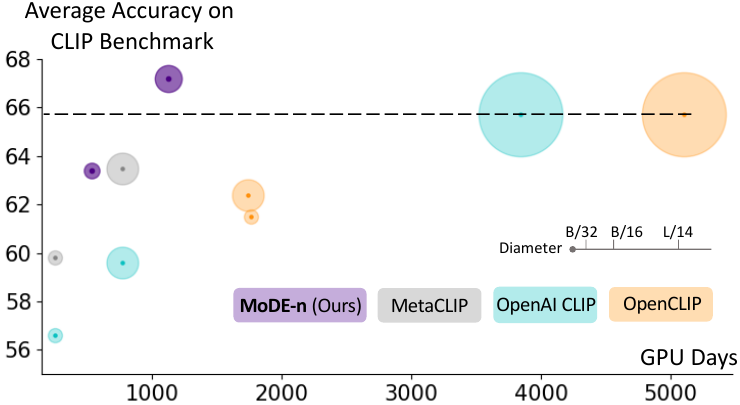} 
    \caption{Summary of average accuracy on CLIP benchmark and pretraining cost (GPU-Hours). The diameter is proportional to the model size, different approaches are color-coded.}
    \label{fig:scaling}
    \vspace{-0.2em}
\end{figure}

Notably, \approach{} provides \emph{an efficient and scalable approach to consume large-scale data without a large batch size that requires more GPUs} (384 Nvidia A100 GPUs) as in OpenCLIP.
As shown in \cref{tab:clip2b5}, based on ViT-B/16 with a batch size of 32K, the \approach{-2} with two data expert models is on par with the ViT-L/14 model by OpenCLIP~\cite{schuhmann2022laion}, while 4 data expert models can outperform the ViT-L/14 by 1.5\% on CLIP benchmark dataset. 
Nevertheless, \approach{} requires much less pretraining cost. As summarized in \cref{fig:scaling}, \approach{-4} ViT-B/16 only requires less-than-35\% of GPU-Hours used for OpenAI CLIP ViT-L/14. 
Compared with OpenCLIP trained on LAION-2B data, \approach{-8} ViT-B/32 data experts can even outperform a single ViT-B/16 model by OpenCLIP but only use 31\% of its GPU-Hours.
In this way, our approach demonstrates great potential for efficient CLIP pretraining with limited GPUs in future.

\bitem{Zero-Shot Robustness}. In addition, to show a consistent gain on different tasks in the CLIP benchmark, we further validate the benefits towards robustness of \approach{} in variants of ImageNet zero-shot classification. As summarized in \cref{tab:robustness}, though there are systematic gaps across variants of ImageNet, learning a set of data experts can improve the zero-shot accuracy on all five variants over the \baseline{} for all model scales, and increasing the number of data experts can still introduce consistent gain. For the accuracies on IN-A and IN-O, the gap between baseline and other approaches is mitigated clearly by \approach{}.
Finally, \approach{-4} achieves the highest average accuracy of all dataset variants among all compared methods.

\bitem{Zero-Shot Retrieval.} We follow OpenCLIP~\cite{schuhmann2022laion} and reports the image/text retrieval results on COCO~\cite{lin2014microsoft} and Flickr30k~\cite{young2014image}. The compared models are trained on billion-scale datasets. As shown in \cref{tab:retrieval}, learning data experts can improve the scores consistently across all model sizes, on COCO, in particular, +3.3\% and +2.7\% in R@1 for image-to-text and text-to-image retrieval respectively by ViT-B/16 models, and we achieve the best performance. For the performance gap between \baseline{} and OpenCLIP, \eg, text retrieval on Flickr30k by ViT-B/32 models, the gap can also be mitigated clearly.

\section{Discussion}\label{sec:discussion}

We first analyze the importance of clustering (\cref{sec:cluster-effect}) and then study the \approach{} design (\cref{sec:cluster-strategy,sec:embedding_type}). Finally, we investigate the potential of our approach in other important research directions (\cref{sec:vision-only,sec:cluster-rank}).

\subsection{Effectiveness of Clustering}\label{sec:cluster-effect} 
As \approach{} ensembles the data experts learned from different clusters, we are first interested in the effects of clustering and consider two variants for ablation.

\begin{table*}[t]
\centering
\setlength\tabcolsep{1.0pt}
\vspace{-0.8em}
\resizebox{\linewidth}{!}
{\renewcommand{\arraystretch}{1.0}
\begin{tabular}{cc|cccccccccccccccccccccccccc}
& \datatag{Average}
& \datatag{ImageNet}
& \datatag{Food-101} 
& \datatag{CIFAR10} 
& \datatag{CIFAR100} 
& \datatag{CUB}
& \datatag{SUN397}
& \datatag{Cars}
& \datatag{Aircraft}
& \datatag{DTD}
& \datatag{Pets}
& \datatag{Caltech-101}
& \datatag{Flowers}
& \datatag{MNIST}
& \datatag{FER-2013}
& \datatag{STL-10}
& \datatag{EuroSAT}
& \datatag{RESISC45}
& \datatag{GTSRB}
& \datatag{KITTI}
& \datatag{Country211}
& \datatag{PCAM}
& \datatag{UCF101}
& \datatag{Kinetics700}
& \datatag{CLEVR}
& \datatag{HatefulMemes}
& \datatag{SST2}\\
    \hlineB{3}
    \rowcolor{lightgray} \multicolumn{28}{l}{400M Image-Caption Pairs}   \\
    \baselinetable{} & 58.2 & 65.5 & 80.6 & {91.3} & 70.2 & 63.4 & 63.0 & 70.7 & 26.8 & 52.8 & 88.7 & 91.9 & 68.5 & 41.5 & \better{35.9} & 95.4 & 52.6 & 64.2 & \better{35.8} & 30.7 & 17.2 & 55.5 & 66.1 & 45.4 & \better{30.6} & 56.4 & 53.4\\ 
    Random-2 & 57.7 & 64.9 & 80.7 & \better{91.4} & 69.6 & 59.8 & 63.0 & \better{72.3} & \better{28.3} & 52.3 & 88.7 & 91.9 & \better{69.4} & 38.1 & 30.8 & 95.4 & \better{52.9} & 62.9 & 33.2 & \better{36.1} & \better{17.3} & 54.4 & 65.7 & 44.7 & 27.1 & 56.2 & 53.0 \\
    Full-2 & 58.3 & 65.9 & 81.0 & 91.2 & 69.9 & 63.8 & \better{63.3} & 71.0 & 27.3 & 52.3 & 88.9 & 91.8 & 69.2 & 42.9 & 33.3 & 95.4 & 52.5 & \better{64.6} & \better{35.8} & 31.2 & 17.0 & \better{56.1} & \better{67.0} & \better{45.5} & 28.7 & \better{57.5} & 53.5 \\
    \approach{-2} & \better{58.6} & \better{66.1} & \better{81.2} & 90.9 & \better{70.5} & \better{65.2} & 63.0 & 72.0 & \better{28.3} & \better{53.5} & \better{89.4} & \cbetter{92.3} & 68.2 & \better{45.2} & 33.5 & 95.4 & 51.9 & 63.7 & 34.9 & 34.2 & \cbetter{17.3} & 54.3 & 65.9 & \better{45.5} & 29.3 & 56.6 & \better{54.6} \\ \shline
    \rowcolor{lightgray} \multicolumn{28}{l}{2.5B Image-Caption Pairs} \\
    \baselinetable{} & 59.8 & 67.6 & 82.6 & 95.2 & 77.7 & 67.8 & 66.8 & 77.2 & 26.9 & 58.9 & 90.9 & 92.5 & 69.7 & 42.7 & 48.3 & 96.3 & 49.9 & 66.5 & 39.2 & 29.3 & 17.7 & 50.0 & 68.0 & 47.6 & 19.4 & 53.5 & 53.1\\ 
    Random-2 & 60.0 & 67.4 & 82.4 & 95.0 & 77.8 & 68.1 & 66.6 & 77.0 & 26.5 & 58.3 & \better{91.0} & 92.3 & 69.0 & \better{45.4} & 47.8 & 96.2 & 50.4 & 66.2 & 43.8 & 30.0 & 17.7 & 50.0 & 67.8 & 47.4 & 20.2 & 53.8 & 52.1 \\
    Full-2 &  60.0 & 67.8 & 82.6 & 95.2 & 77.7 & 68.4 & 66.7 & 77.7 & 27.7 & 58.6 & 90.9 & 92.5 & 69.9 & 43.6 & 48.7 & \better{96.4} & 50.1 & 66.0 & 41.7 & 28.2 & 17.9 & 50.0 & 68.4 & \better{47.7} & 19.3 & \better{53.9} & 52.8 \\
    \approach{-2} & \better{61.2} & \better{68.7} & \better{84.1} & \better{95.3} & \better{78.6} & \better{69.5} & \better{67.0} & \better{80.8} & \cbetter{30.9} & \better{60.6} & \better{91.0} & \better{92.9} & \better{71.9} & 40.8 &\better{50.4} & 96.3 & \better{51.3} & \cbetter{67.9} & \better{44.2} & \better{31.4} & \better{18.3} & \better{51.3} & \cbetter{69.0} & 47.4 & \better{23.2} & 52.6 & \better{54.4} \\ 
    \hlineB{3}
    \end{tabular}}
    \caption{Ablation Study for performance gain via Clustering by VIT-B/32.}\label{tab:dual-ablate}
\end{table*}

\begin{table}[t]
    \vspace{-1.em}
    \centering
    \footnotesize
    \resizebox{\linewidth}{!}{
    \renewcommand{\arraystretch}{1.05}
    \begin{tabular}{c|cc|cc}
    \hlineB{3}
    Approach      & CLIP Avg.    & ImageNet     & CLIP Avg.      & ImageNet      \\ \hline
    \baselinetable{}          & 58.2      & 65.6         & 59.8       & 67.7          \\
    OneStep-2            & 58.0      & 65.0         & 59.8     & 67.6         \\
    CoarseCluster-2 & 58.5      & 66.1         & 60.6       & 68.6          \\
    \approach{-2}   & \better{58.6}      & \better{66.1}         & \better{61.2}       & \better{68.7}          \\ \hline \hline
    CoarseCluster-4 & 58.7      & 66.2         & 61.3       & 68.5          \\
    \approach{-4}   & \better{59.0}      & \better{66.4}         & \better{61.7}       & \better{68.8}          \\
    \hlineB{3}
    \rowcolor{lightgray} Pre-Train Dataset& \multicolumn{2}{c|}{400M Image-Caption Pairs} & \multicolumn{2}{c}{2.5B Image-Caption Pairs} \\ \hlineB{3}
    \end{tabular}}
    \caption{Ablation study for Clustering Strategy by ViT-B/32.}
    \label{tab:ensemble-ablate}
\end{table}

Though model ensembling~\cite{jordan2015machine} can provide gains over a single model, we are interested in how a naive ensembling of models trained on similar distribution performs compared to MoDE with data specialization. In \cref{tab:dual-ablate},
we train two ViT-B/32 CLIP models on the same training data without clustering, and then average the model outputs for prediction (Full-2). 
This achieves a similar performance as the baseline.
Thus, the clustering is essential for \approach{}.  

Furthermore, we randomly split the training data into two subsets, and specialize a data expert for each subset (Random-2). 
For a fair comparison, we mimic the size of subsets by \approach{-2} in the random splitting, and all data experts use the same seed model.
As the data split is not obtained through clustering, we still only use the average of model outputs for evaluation. However, though Random-2 can provide small improvement when trained on 2.5B image-caption pairs (60.0 \vs 59.8), there is a noticeable drop when training on the 400M pairs (57.7 \vs 58.2).

\begin{figure*}{}
  \begin{minipage}[b]{0.33\textwidth}
    \centering
    \includegraphics[width=\textwidth]{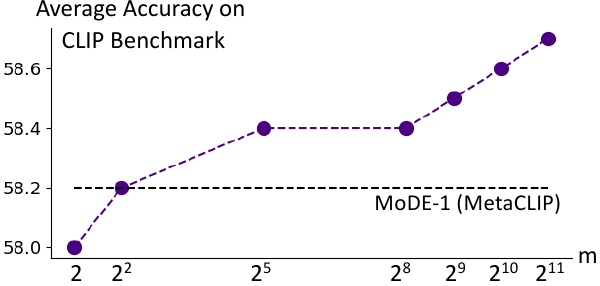}
    \captionof{figure}{Ablation on \# of clusters in Step 1.}
    \label{fig:specturm-fine}
  \end{minipage}
  \hfill
  \begin{minipage}[b]{0.38\textwidth}
    \centering
    \resizebox{\linewidth}{!}
    {\renewcommand{\arraystretch}{1.2}
        \begin{tabular}{cccccc}
            \hlineB{3}
            Modality &  Model  & CLIP Eval.  &  ImageNet  \\
            \hline
            Image & DINOv2 & 58.1 & 65.2 \\
            Image & CLIP Seed & 58.3 & 64.7 \\
            Image \& Lang. & CLIP Seed & 58.4 & 65.5 \\ 
            Lang. & CLIP Seed & 58.3 & 65.4 \\\hline
            Lang. & SimCSE~\cite{gao2021simcse} & \better{58.6} & \better{66.1} \\
            \hlineB{3}
        \end{tabular}
    }
    \captionof{table}{Ablation Study on Embedding Types.}
    \label{tab:ablate-embedding}
    \end{minipage}
  \hfill
  \begin{minipage}[b]{0.28\textwidth}
    \centering
    \renewcommand{\arraystretch}{1.4}
    \resizebox{\textwidth}{!}
    {
        \begin{tabular}{c|cccc}
            \hlineB{3}
            Approach & B/32  & B/16  & L/14  &  G/14 \\ 
            \hline
            OpenAI CLIP     & 63.3    & 68.4       & 75.6     & -   \\
            OpenCLIP & 66.6 & 70.2  & 75.3    & 80.1  \\
            \baselinetable{} & 67.6 & 72.1     & 79.2   & -   \\
            Ours &  \better{71.4} & \better{75.3} & \better{80.3} & - \\ \hlineB{3}
        \end{tabular}
    }
    \captionof{table}{ImageNet Zero-shot Accuracy via Prior-based Data Ranking.}
    \label{tab:retrieval-enhance}
    \end{minipage}
\end{figure*}

\subsection{Clustering Strategy}\label{sec:cluster-strategy} 

Instead of obtaining the data clusters in a single step, \approach{} employs a two-step clustering strategy to discover the centers of fine-grained cluster $S$, which are used to properly model the correlation between task metadata and the conditions (\cref{sec:clustering}). We provide ablation studies below to demonstrate this necessity for model ensembling. 

Firstly, we evaluate the one-step clustering alternative, \ie, $m=n$, and for simplicity, we only learn two data experts (OneStep-2) based on ViT-B/32. As shown in \cref{tab:ensemble-ablate}, we summarize the average score on the CLIP benchmark and stand out the accuracy of ImageNet as it has the most number of classes. As the cluster centers are not representative enough to model the correlation with task metadata, model ensembling in OneStep-2 can even result in a slight drop. We do observe that each data expert alone can outperform \baseline{} on different tasks in the CLIP benchmark but it is difficult to pick  correctly. 

Then, we follow the two-step clustering but alter the number of fine-grained clusters $m$ in the first step. As plotted in \cref{fig:specturm-fine}, we summarize the results of \approach{-2} trained on 400M image-caption pairs. With increasing $m$, we observed that the average accuracy on the CLIP evaluation benchmark improves consistently. Though the performance can be improved slightly when $m$ is increased from 1024 to 2048, the computational cost during data clustering is also higher. We set $m=1024$ in the main experiments.

Lastly, as another piece of evidence, we keep $m$ as 1024 but use the coarse-grained cluster centers in Step 2, to determine the ensembling weights (CoarseCluster). As shown in \cref{tab:ensemble-ablate} , as the meta clusters are not representative enough to obtain good ensembling weight, the resulting accuracy improvement is trivial. When we increase the number of data experts from 2 to 4, the gap between CoarseCluster-4 and \approach{-4} is even enlarged, which further demonstrates the importance of using fine-grained clusters to determine the ensembling weight for data experts in our \approach{}.

\subsection{Embeddings for Clustering}\label{sec:embedding_type}
We further validate the importance of using language embeddings.
In addition to SimCSE~\cite{gao2021simcse} language embedding, we investigate the following embeddings for clustering: (1) image embedding from the open-sourced DINOv2~\cite{oquab2023dinov2}; (2) image and/or text embeddings from the seed model (\ie, the partially trained MetaCLIP checkpoints on the 27th epoch). 
When the image embeddings are used for clustering, for each test image, we use its similarity with all fine-grained cluster centers to determine the logits ensemble weights. When both image and text embeddings are used, we use their concatenation as the feature for clustering.
Without loss of generality, we compare with \approach{-2} trained on 400M pairs and set $m=1024$ for fair comparison. We summarize the scores in \cref{tab:ablate-embedding} and report the zero-shot accuracy CLIP benchmark and ImageNet.

Firstly, by using image embeddings for clustering, the resulting models underperform MetaCLIP, in particular on ImageNet, and we believe the main reason is that the image embedding contains low-level details.
As such, the cluster centers are not representative of model ensembling. 

\begin{figure}[!htb]
    \centering
    \includegraphics[width=0.95\linewidth]{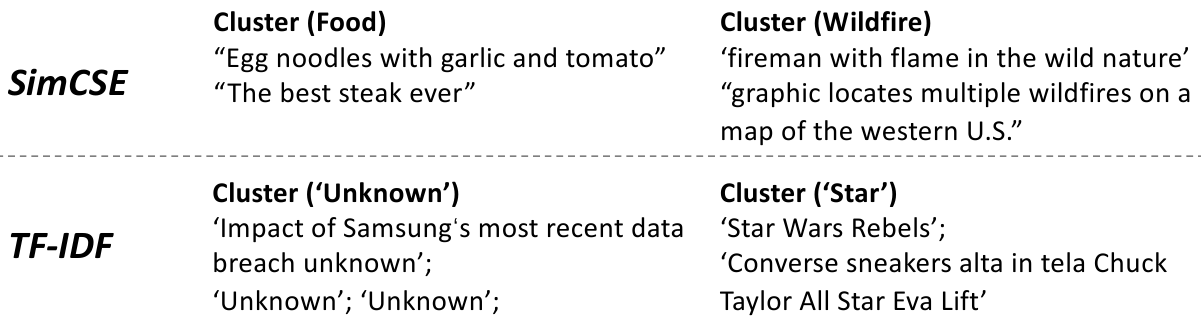}
    \caption{Representative instances for each cluster.}
    \label{fig:cluster_visualization}
\end{figure}

Furthermore, utilizing the language embeddings from the seed model yields only marginal performance improvement. This suggests that the CLIP embedding may still fall short of discerning high-level semantic correlations. 
This occurs as the language embeddings are influenced by image embeddings, potentially overlooking high-level semantics not depicted in corresponding images. For example, abstract concepts such as ``travel'', ``product'', and ``politics'' may lack corresponding visual elements.
In contrast, the SimCSE text embeddings pretrained on large text corpora can understand abstract concepts for clustering. 
As another evidence, we compare the clustering based on language embeddings and use TF-IDF embeddings as reference. As the TF-IDF embedding is determined by on the frequency of words, the clusters on TF-IDF embeddings shown in \cref{fig:cluster_visualization} can only group captions with the same words, and struggle to provide abstract concepts via discrete text tokens. In contrast, using SimCSE embeddings can group the captions with coherent semantics (\eg, food). 

\subsection{Application of Vision Encoders}\label{sec:vision-only}

Besides zero-shot generalization, the set of vision encoders can also be directly ensembled in downstream application.
Notably, all vision encoders are equally weighted, and we do not need any cluster center (\ie, cluster-independent), which is generalizable to the case where the language metadata such as class names is not available.

Firstly, we ensemble the encoder outputs and use ImageNet classification for evaluation. Specifically, for each image, we concatenate the outputs from all ($n$) vision encoders as the representation and feed it into a linear layer for classification. To maintain reasonable training cost, only linear probing is considered where we exclusively train the linear classifier from scratch and fix all vision encoders. As shown in \cref{tab:linear_sum}, our \approach{} achieves consistent and clear performance gain over \baseline{}.

\begin{table}[!htb]
    \centering
    \setlength\tabcolsep{1.0pt}
    \resizebox{0.9\linewidth}{!}{
    \begin{tabular}{c|ccc|ccc}
    \shline
    \multirow{2}{*}{Model}    & \multicolumn{3}{c|}{Linear Probe$^*$}    & \multicolumn{3}{c}{Linear Probe} \\
             & ~~~B/32~~~ & ~~~B/16~~~ & ~~~L/14~~~ & ~~~B/32~~~ & ~~~B/16~~~ & ~~~L/14~~~ \\ \hline
    \baselinetable{} & \better{69.3} & 73.3 & 80.3 & 67.5 & 73.8 & 82.3      \\
    MoDE-2          & 68.9 & 73.8 & 80.6 & 71.3 & 76.9 & 83.9      \\
    MoDE-4          & 69.1 & \better{74.5} & \better{80.7} & \better{74.1} & \better{79.6} & \better{84.7}    \\ \shline
    \multicolumn{7}{l}{\small $^*$: Initialize classifier with language embeddings as in OpenCLIP~\cite{schuhmann2022laion}.}
    \end{tabular}}
    \caption{Performance comparison on ImageNet via linear probing.}
    \label{tab:linear_sum}
\end{table}

As shown in \cref{tab:zsl_linear}, we evaluate all vision encoders by \approach{-4} ViT-B/16 independently and report the accuracy via linear probing and finetuning (\ie, all parameters are trained). Linear probing on the concatenated features achieves higher score than finetuning a single model (79.6 Vs. 76.7) but with much less training cost. 

\begin{table}[!htb]
    \centering
    \setlength\tabcolsep{1.0pt}
    \resizebox{0.95\linewidth}{!}{
    \begin{tabular}{c|cccc}
    \shline
    Data Experts & ~~~Zero-Shot~~~ & ~~~Linear Probe$^*$~~~ & ~~~Linear Probe~~~ & ~~~Finetune~~~ \\ \hline
    \baselinetable{}	& 72.1	& 73.3	& 73.8	& 76.7 \\
    0            & 63.3      & 66.4                 & 67.3         & 75.7     \\
    1            & 68.5      & 71.3                 & 72.0         & 76.9     \\
    2            & 65.2      & 68.2                 & 68.8         & 76.3     \\
    3            & \better{72.9 }     & \better{74.9}                 & \better{74.2}         & \better{77.2}     \\
    \shline
    \multicolumn{5}{l}{\small $^*$: Initialize classifier with language embeddings as in OpenCLIP~\cite{schuhmann2022laion}.}
    \end{tabular}}
    \caption{Evaluation for each data expert in \approach{-4} ViT-B/16.}
    \label{tab:zsl_linear}
\end{table}

In addition, by comparing among vision encoders, the data expert achieving higher zero-shot accuracy also hits the best score in both linear probing and finetuning, indicating a consistent correlation benefited from the strong encoder initialization. In this way, by training data expert on each coarse-grained cluster, we increase the quality negative within each mini-batch to learn stronger vision encoders effectively.
Finally, the parameters can also be averaged and then used as initialization of a single network for finetuning, and more details can be found in the Supp.

\subsection{Training Priority of Data Experts}\label{sec:cluster-rank}

As the data experts can be trained asynchronously, \approach{} introduces flexibility in the data expert training priority. Below we demonstrate the robustness and effectiveness of \approach{} when the data experts are trained in order. 

\begin{figure}[t]
    \centering
    \includegraphics[width=0.95\linewidth]{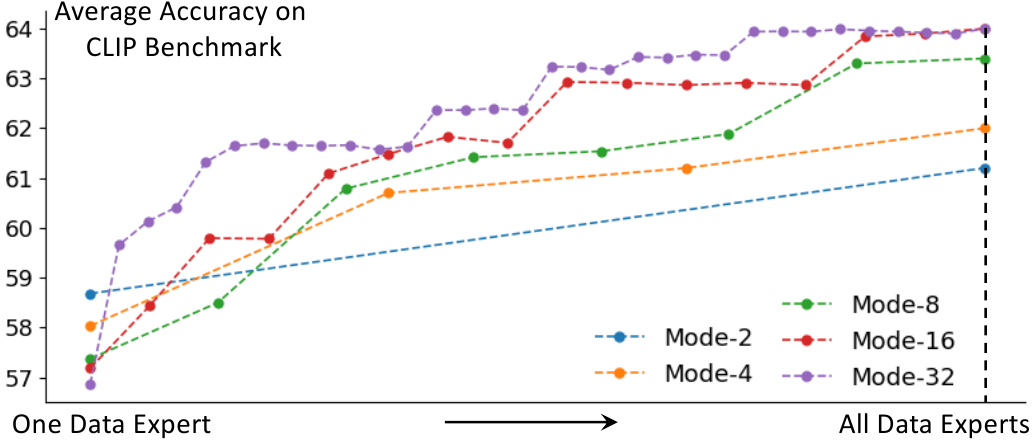}
    \caption{CLIP benchmark accuracy by \approach{}-$n$ when the data experts based on ViT-B/32 are developed in order and added to the system progressively. The pre-train set contains 2.5B pairs.}
    \label{fig:continual}
\end{figure}

Firstly, we rank the conditions, \ie, coarse-level clusters, to determine the training priority of data experts. This is useful when the computational resource is not sufficient to learn a giant dense model or all data experts together. We use the diversity of fine-grained clusters as a reference, and first train the model on the condition with the largest range, \ie, the average distance between fine-grained clusters and the coarse-grained center. As shown in \cref{fig:continual}, we vary the total number of ViT-B/32 data experts, \ie, $n$, from 2 to 32 and summarize the average accuracy on the CLIP benchmark. When the data experts are gradually included, the performance keeps increasing.

In this way, instead of learning from all data simultaneously, \approach enables progressive integration of new data experts, enabling dynamic updates. \approach holds promise for applications such as \emph{online} and {continual learning}. With each new set of data, it has the flexibility to update a pre-trained data expert, or to learn a new data expert. This is particularly valuable when the incoming data are unprecedented to the existing expert system. We leave the trade-off between catastrophic forgetting~\cite{forget} and effective adaption as the futrure work.

At the same time, we can also select the clusters given the task metadata as prior following the retrieval-enhanced setup~\cite{iscen2023retrieval}. 
When the metadata is accessible, we use the SimCSE~\cite{gao2021simcse} to extract their embeddings and retrieve the nearest fine-grained clusters for each of them. Then, the data expert trained on the selected clusters is of highest training priority, and we only train that single data expert for evaluation while the rest clusters can be left for future continual MoDE pretraining if needed.
We take ImageNet as an example where the 1000 class names are used to retrieve clusters. As summarized in \cref{tab:retrieval-enhance}, adapting our approach can improve the efficiency of network training significantly and can even escalate the performance along the model scale in most cases. For example, our ViT-B/16 outperforms the L/14 models by OpenAI CLIP/ OpenCLIP and our ViT-L/14 even outperforms the ViT-G/14 in OpenCLIP. Besides, as detailed in Suppl., \approach{} can also be aligned for a set of downstream tasks, \eg, CLIP benchmarks.

In summary, \approach can be applied to different types of downstream tasks. Meanwhile, the coarse-level clustering in the second step tentatively assumes the fine-grained clusters should be split into disjoint groups with overlap. 
We believe the fine-grained clusters can also be grouped flexibly and we leave it for future work.

\section{Conclusion}

The success of CLIP depends on the quality \textit{negative} samples. As the \emph{false negative} noise in web-crawled pairs hurts training effectiveness, scaling CLIP on large-scale data presents unique challenges in terms of training efficiency and computational bottlenecks.
To this end, we have presented Mixture of Data Experts (\approach{}) to asynchronously train a group of \emph{data experts}. Each expert model is trained on a set of fine-grained clusters where the data in each cluster is of coherent semantics and all data experts are trained individually.
During inference, the outputs are selectively ensembled based on the requirements for each task and modeled by the correlation between task metadata and fine-grained cluster centers.
Empirically, \approach{} significantly outperforms  OpenCLIP and OpenAI CLIP on standard benchmarks with less than 35\% training cost.
Furthermore, the image embeddings extracted by all data experts can be combined to enhance the representation of visual information.
We plan to adapt MoDE for generative models in the future.

\section*{Acknowledgement}
\noindent The authors would like to thank Xinlei Chen and Margaret Li for constructive discussion.


{\small
\bibliographystyle{ieee_fullname}
\bibliography{reference}

\begin{thebibliography}{10}\itemsep=-1pt

\bibitem{blahut2010fast}
Richard~E Blahut.
\newblock {\em Fast algorithms for signal processing}.
\newblock Cambridge University Press, 2010.

\bibitem{chen2020simple}
Ting Chen, Simon Kornblith, Mohammad Norouzi, and Geoffrey Hinton.
\newblock A simple framework for contrastive learning of visual representations.
\newblock In {\em International conference on machine learning}, pages 1597--1607. PMLR, 2020.

\bibitem{deng2009imagenet}
Jia Deng, Wei Dong, Richard Socher, Li-Jia Li, Kai Li, and Li Fei-Fei.
\newblock Imagenet: A large-scale hierarchical image database.
\newblock In {\em 2009 IEEE conference on computer vision and pattern recognition}, pages 248--255. Ieee, 2009.

\bibitem{desai2023hyperbolic}
Karan Desai, Maximilian Nickel, Tanmay Rajpurohit, Justin Johnson, and Shanmukha~Ramakrishna Vedantam.
\newblock Hyperbolic image-text representations.
\newblock In {\em International Conference on Machine Learning}, pages 7694--7731. PMLR, 2023.

\bibitem{dhillon2001concept}
Inderjit~S Dhillon and Dharmendra~S Modha.
\newblock Concept decompositions for large sparse text data using clustering.
\newblock {\em Machine learning}, 42:143--175, 2001.

\bibitem{dosovitskiy2020image}
Alexey Dosovitskiy, Lucas Beyer, Alexander Kolesnikov, Dirk Weissenborn, Xiaohua Zhai, Thomas Unterthiner, Mostafa Dehghani, Matthias Minderer, Georg Heigold, Sylvain Gelly, et~al.
\newblock An image is worth 16x16 words: Transformers for image recognition at scale.
\newblock {\em arXiv preprint arXiv:2010.11929}, 2020.

\bibitem{eigen2013learning}
David Eigen, Marc'Aurelio Ranzato, and Ilya Sutskever.
\newblock Learning factored representations in a deep mixture of experts.
\newblock {\em arXiv preprint arXiv:1312.4314}, 2013.

\bibitem{vsepp}
Fartash Faghri, David~J. Fleet, Jamie~Ryan Kiros, and Sanja Fidler.
\newblock {VSE++:} improving visual-semantic embeddings with hard negatives.
\newblock In {\em British Machine Vision Conference 2018, {BMVC} 2018, Newcastle, UK, September 3-6, 2018}, page~12. {BMVA} Press, 2018.

\bibitem{fedus2022switch}
William Fedus, Barret Zoph, and Noam Shazeer.
\newblock Switch transformers: Scaling to trillion parameter models with simple and efficient sparsity.
\newblock {\em The Journal of Machine Learning Research}, 23(1):5232--5270, 2022.

\bibitem{gammerman2013learning}
Alex Gammerman, Volodya Vovk, and Vladimir Vapnik.
\newblock Learning by transduction.
\newblock {\em arXiv preprint arXiv:1301.7375}, 2013.

\bibitem{gandelsman2022test}
Yossi Gandelsman, Yu Sun, Xinlei Chen, and Alexei Efros.
\newblock Test-time training with masked autoencoders.
\newblock {\em Advances in Neural Information Processing Systems}, 35:29374--29385, 2022.

\bibitem{gao2021simcse}
Tianyu Gao, Xingcheng Yao, and Danqi Chen.
\newblock Simcse: Simple contrastive learning of sentence embeddings.
\newblock In {\em 2021 Conference on Empirical Methods in Natural Language Processing, EMNLP 2021}, pages 6894--6910. Association for Computational Linguistics (ACL), 2021.

\bibitem{gururangan2023scaling}
Suchin Gururangan, Margaret Li, Mike Lewis, Weijia Shi, Tim Althoff, Noah~A Smith, and Luke Zettlemoyer.
\newblock Scaling expert language models with unsupervised domain discovery.
\newblock {\em arXiv preprint arXiv:2303.14177}, 2023.

\bibitem{pmlr-v9-gutmann10a}
Michael Gutmann and Aapo Hyvärinen.
\newblock Noise-contrastive estimation: A new estimation principle for unnormalized statistical models.
\newblock In Yee~Whye Teh and Mike Titterington, editors, {\em Proceedings of the Thirteenth International Conference on Artificial Intelligence and Statistics}, volume~9 of {\em Proceedings of Machine Learning Research}, pages 297--304, Chia Laguna Resort, Sardinia, Italy, 13--15 May 2010. PMLR.

\bibitem{han2022few}
Guangxing Han, Jiawei Ma, Shiyuan Huang, Long Chen, and Shih-Fu Chang.
\newblock Few-shot object detection with fully cross-transformer.
\newblock In {\em Proceedings of the IEEE/CVF conference on computer vision and pattern recognition}, pages 5321--5330, 2022.

\bibitem{he2020momentum}
Kaiming He, Haoqi Fan, Yuxin Wu, Saining Xie, and Ross Girshick.
\newblock Momentum contrast for unsupervised visual representation learning.
\newblock In {\em Proceedings of the IEEE/CVF conference on computer vision and pattern recognition}, pages 9729--9738, 2020.

\bibitem{iscen2023retrieval}
Ahmet Iscen, Mathilde Caron, Alireza Fathi, and Cordelia Schmid.
\newblock Retrieval-enhanced contrastive vision-text models.
\newblock {\em arXiv preprint arXiv:2306.07196}, 2023.

\bibitem{jacobs1991adaptive}
Robert~A Jacobs, Michael~I Jordan, Steven~J Nowlan, and Geoffrey~E Hinton.
\newblock Adaptive mixtures of local experts.
\newblock {\em Neural computation}, 3(1):79--87, 1991.

\bibitem{jia2021scaling}
Chao Jia, Yinfei Yang, Ye Xia, Yi-Ting Chen, Zarana Parekh, Hieu Pham, Quoc Le, Yun-Hsuan Sung, Zhen Li, and Tom Duerig.
\newblock Scaling up visual and vision-language representation learning with noisy text supervision.
\newblock In {\em International conference on machine learning}, pages 4904--4916. PMLR, 2021.

\bibitem{johnson2019billion}
Jeff Johnson, Matthijs Douze, and Herv{\'e} J{\'e}gou.
\newblock Billion-scale similarity search with {GPUs}.
\newblock {\em IEEE Transactions on Big Data}, 7(3):535--547, 2019.

\bibitem{jordan1994hierarchical}
Michael~I Jordan and Robert~A Jacobs.
\newblock Hierarchical mixtures of experts and the em algorithm.
\newblock {\em Neural computation}, 6(2):181--214, 1994.

\bibitem{jordan2015machine}
Michael~I Jordan and Tom~M Mitchell.
\newblock Machine learning: Trends, perspectives, and prospects.
\newblock {\em Science}, 349(6245):255--260, 2015.

\bibitem{kalantidis2020hard}
Yannis Kalantidis, Mert~Bulent Sariyildiz, Noe Pion, Philippe Weinzaepfel, and Diane Larlus.
\newblock Hard negative mixing for contrastive learning.
\newblock {\em Advances in Neural Information Processing Systems}, 33:21798--21809, 2020.

\bibitem{forget}
James Kirkpatrick, Razvan Pascanu, Neil~C. Rabinowitz, Joel Veness, Guillaume Desjardins, Andrei~A. Rusu, Kieran Milan, John Quan, Tiago Ramalho, Agnieszka Grabska{-}Barwinska, Demis Hassabis, Claudia Clopath, Dharshan Kumaran, and Raia Hadsell.
\newblock Overcoming catastrophic forgetting in neural networks.
\newblock {\em CoRR}, abs/1612.00796, 2016.

\bibitem{lepikhin2020gshard}
Dmitry Lepikhin, HyoukJoong Lee, Yuanzhong Xu, Dehao Chen, Orhan Firat, Yanping Huang, Maxim Krikun, Noam Shazeer, and Zhifeng Chen.
\newblock Gshard: Scaling giant models with conditional computation and automatic sharding.
\newblock {\em arXiv preprint arXiv:2006.16668}, 2020.

\bibitem{li2022branch}
Margaret Li, Suchin Gururangan, Tim Dettmers, Mike Lewis, Tim Althoff, Noah~A Smith, and Luke Zettlemoyer.
\newblock Branch-train-merge: Embarrassingly parallel training of expert language models.
\newblock {\em arXiv preprint arXiv:2208.03306}, 2022.

\bibitem{clipa}
Xianhang Li, Zeyu Wang, and Cihang Xie.
\newblock An inverse scaling law for clip training.
\newblock {\em arXiv preprint arXiv:2305.07017}, 2023.

\bibitem{li2023scaling}
Yanghao Li, Haoqi Fan, Ronghang Hu, Christoph Feichtenhofer, and Kaiming He.
\newblock Scaling language-image pre-training via masking.
\newblock In {\em Proceedings of the IEEE/CVF Conference on Computer Vision and Pattern Recognition}, pages 23390--23400, 2023.

\bibitem{lin2014microsoft}
Tsung-Yi Lin, Michael Maire, Serge Belongie, James Hays, Pietro Perona, Deva Ramanan, Piotr Doll{\'a}r, and C~Lawrence Zitnick.
\newblock Microsoft coco: Common objects in context.
\newblock In {\em Computer Vision--ECCV 2014: 13th European Conference, Zurich, Switzerland, September 6-12, 2014, Proceedings, Part V 13}, pages 740--755. Springer, 2014.

\bibitem{liu2023learning}
Haotian Liu, Kilho Son, Jianwei Yang, Ce Liu, Jianfeng Gao, Yong~Jae Lee, and Chunyuan Li.
\newblock Learning customized visual models with retrieval-augmented knowledge.
\newblock In {\em Proceedings of the IEEE/CVF Conference on Computer Vision and Pattern Recognition}, pages 15148--15158, 2023.

\bibitem{ma2021partner}
Jiawei Ma, Hanchen Xie, Guangxing Han, Shih-Fu Chang, Aram Galstyan, and Wael Abd-Almageed.
\newblock Partner-assisted learning for few-shot image classification.
\newblock In {\em Proceedings of the IEEE/CVF International Conference on Computer Vision}, pages 10573--10582, 2021.

\bibitem{malinen2014balanced}
Mikko~I Malinen and Pasi Fr{\"a}nti.
\newblock Balanced k-means for clustering.
\newblock In {\em Structural, Syntactic, and Statistical Pattern Recognition: Joint IAPR International Workshop, S+ SSPR 2014, Joensuu, Finland, August 20-22, 2014. Proceedings}, pages 32--41. Springer, 2014.

\bibitem{mitchell1997machine}
Tom~M Mitchell.
\newblock Machine learning, 1997.

\bibitem{mu2022slip}
Norman Mu, Alexander Kirillov, David Wagner, and Saining Xie.
\newblock Slip: Self-supervision meets language-image pre-training.
\newblock In {\em European Conference on Computer Vision}, pages 529--544. Springer, 2022.

\bibitem{mustafa2022multimodal}
Basil Mustafa, Carlos Riquelme, Joan Puigcerver, Rodolphe Jenatton, and Neil Houlsby.
\newblock Multimodal contrastive learning with limoe: the language-image mixture of experts.
\newblock {\em Advances in Neural Information Processing Systems}, 35:9564--9576, 2022.

\bibitem{oquab2023dinov2}
Maxime Oquab, Timoth{\'e}e Darcet, Th{\'e}o Moutakanni, Huy Vo, Marc Szafraniec, Vasil Khalidov, Pierre Fernandez, Daniel Haziza, Francisco Massa, Alaaeldin El-Nouby, et~al.
\newblock Dinov2: Learning robust visual features without supervision.
\newblock {\em arXiv preprint arXiv:2304.07193}, 2023.

\bibitem{supportset}
Mandela Patrick, Po{-}Yao Huang, Yuki~Markus Asano, Florian Metze, Alexander~G. Hauptmann, Jo{\~{a}}o~F. Henriques, and Andrea Vedaldi.
\newblock Support-set bottlenecks for video-text representation learning.
\newblock In {\em 9th International Conference on Learning Representations, {ICLR} 2021, Virtual Event, Austria, May 3-7, 2021}. OpenReview.net, 2021.

\bibitem{pham2023combined}
Hieu Pham, Zihang Dai, Golnaz Ghiasi, Kenji Kawaguchi, Hanxiao Liu, Adams~Wei Yu, Jiahui Yu, Yi-Ting Chen, Minh-Thang Luong, Yonghui Wu, et~al.
\newblock Combined scaling for zero-shot transfer learning.
\newblock {\em Neurocomputing}, 555:126658, 2023.

\bibitem{radford2021learning}
Alec Radford, Jong~Wook Kim, Chris Hallacy, Aditya Ramesh, Gabriel Goh, Sandhini Agarwal, Girish Sastry, Amanda Askell, Pamela Mishkin, Jack Clark, et~al.
\newblock Learning transferable visual models from natural language supervision.
\newblock In {\em International conference on machine learning}, pages 8748--8763. PMLR, 2021.

\bibitem{riquelme2021scaling}
Carlos Riquelme, Joan Puigcerver, Basil Mustafa, Maxim Neumann, Rodolphe Jenatton, Andr{\'e} Susano~Pinto, Daniel Keysers, and Neil Houlsby.
\newblock Scaling vision with sparse mixture of experts.
\newblock {\em Advances in Neural Information Processing Systems}, 34:8583--8595, 2021.

\bibitem{sachidananda2023global}
Vin Sachidananda, Ziyi Yang, and Chenguang Zhu.
\newblock Global selection of contrastive batches via optimization on sample permutations.
\newblock In {\em Proceedings of the 40th International Conference on Machine Learning}, 2023.

\bibitem{sain1996nature}
Stephan~R Sain.
\newblock The nature of statistical learning theory, 1996.

\bibitem{schuhmann2022laion}
Christoph Schuhmann, Romain Beaumont, Richard Vencu, Cade Gordon, Ross Wightman, Mehdi Cherti, Theo Coombes, Aarush Katta, Clayton Mullis, Mitchell Wortsman, et~al.
\newblock Laion-5b: An open large-scale dataset for training next generation image-text models.
\newblock {\em Advances in Neural Information Processing Systems}, 35:25278--25294, 2022.

\bibitem{schuhmann2021laion}
Christoph Schuhmann, Richard Vencu, Romain Beaumont, Robert Kaczmarczyk, Clayton Mullis, Aarush Katta, Theo Coombes, Jenia Jitsev, and Aran Komatsuzaki.
\newblock Laion-400m: Open dataset of clip-filtered 400 million image-text pairs.
\newblock {\em arXiv preprint arXiv:2111.02114}, 2021.

\bibitem{shazeer2016outrageously}
Noam Shazeer, Azalia Mirhoseini, Krzysztof Maziarz, Andy Davis, Quoc Le, Geoffrey Hinton, and Jeff Dean.
\newblock Outrageously large neural networks: The sparsely-gated mixture-of-experts layer.
\newblock In {\em International Conference on Learning Representations}, 2016.

\bibitem{snell2017prototypical}
Jake Snell, Kevin Swersky, and Richard Zemel.
\newblock Prototypical networks for few-shot learning.
\newblock {\em Advances in neural information processing systems}, 30, 2017.

\bibitem{sun2020test}
Yu Sun, Xiaolong Wang, Zhuang Liu, John Miller, Alexei Efros, and Moritz Hardt.
\newblock Test-time training with self-supervision for generalization under distribution shifts.
\newblock In {\em International conference on machine learning}, pages 9229--9248. PMLR, 2020.

\bibitem{xie2023ra}
Chen-Wei Xie, Siyang Sun, Xiong Xiong, Yun Zheng, Deli Zhao, and Jingren Zhou.
\newblock Ra-clip: Retrieval augmented contrastive language-image pre-training.
\newblock In {\em Proceedings of the IEEE/CVF Conference on Computer Vision and Pattern Recognition}, pages 19265--19274, 2023.

\bibitem{videoclip}
Hu Xu, Gargi Ghosh, Po{-}Yao Huang, Dmytro Okhonko, Armen Aghajanyan, Florian Metze, Luke Zettlemoyer, and Christoph Feichtenhofer.
\newblock Videoclip: Contrastive pre-training for zero-shot video-text understanding.
\newblock In Marie{-}Francine Moens, Xuanjing Huang, Lucia Specia, and Scott~Wen{-}tau Yih, editors, {\em Proceedings of the 2021 Conference on Empirical Methods in Natural Language Processing, {EMNLP}, 7-11 November, 2021}, pages 6787--6800. Association for Computational Linguistics, 2021.

\bibitem{xu2023demystifying}
Hu Xu, Saining Xie, Xiaoqing~Ellen Tan, Po-Yao Huang, Russell Howes, Vasu Sharma, Shang-Wen Li, Gargi Ghosh, Luke Zettlemoyer, and Christoph Feichtenhofer.
\newblock Demystifying clip data.
\newblock {\em arXiv preprint arXiv:2309.16671}, 2023.

\bibitem{yang2023tempclr}
Yuncong Yang, Jiawei Ma, Shiyuan Huang, Long Chen, Xudong Lin, Guangxing Han, and Shih-Fu Chang.
\newblock Temp{CLR}: Temporal alignment representation with contrastive learning.
\newblock In {\em The Eleventh International Conference on Learning Representations}, 2023.

\bibitem{young2014image}
Peter Young, Alice Lai, Micah Hodosh, and Julia Hockenmaier.
\newblock From image descriptions to visual denotations: New similarity metrics for semantic inference over event descriptions.
\newblock {\em Transactions of the Association for Computational Linguistics}, 2:67--78, 2014.

\bibitem{yu2022coca}
Jiahui Yu, Zirui Wang, Vijay Vasudevan, Legg Yeung, Mojtaba Seyedhosseini, and Yonghui Wu.
\newblock Coca: Contrastive captioners are image-text foundation models.
\newblock {\em arXiv preprint arXiv:2205.01917}, 2022.

\bibitem{zhai2022lit}
Xiaohua Zhai, Xiao Wang, Basil Mustafa, Andreas Steiner, Daniel Keysers, Alexander Kolesnikov, and Lucas Beyer.
\newblock Lit: Zero-shot transfer with locked-image text tuning.
\newblock In {\em Proceedings of the IEEE/CVF Conference on Computer Vision and Pattern Recognition}, pages 18123--18133, 2022.

\end{thebibliography}
}


\appendix

\begin{table*}[!htb]
\vspace{-1.em}
\centering
\setlength\tabcolsep{1.0pt}
\resizebox{\linewidth}{!}{
\begin{tabular}{lc|cccccccccccccccccccccccccc}
& \datatag{Average}
& \datatag{ImageNet}
& \datatag{Food-101} 
& \datatag{CIFAR10} 
& \datatag{CIFAR100} 
& \datatag{CUB}
& \datatag{SUN397}
& \datatag{Cars}
& \datatag{Aircraft}
& \datatag{DTD}
& \datatag{Pets}
& \datatag{Caltech-101}
& \datatag{Flowers}
& \datatag{MNIST}
& \datatag{FER-2013}
& \datatag{STL-10}
& \datatag{EuroSAT}
& \datatag{RESISC45}
& \datatag{GTSRB}
& \datatag{KITTI}
& \datatag{Country211}
& \datatag{PCAM}
& \datatag{UCF101}
& \datatag{Kinetics700}
& \datatag{CLEVR}
& \datatag{HatefulMemes}
& \datatag{SST2}\\
\shline
MetaCLIP & 59.8 & 67.6 & 82.6 & 95.2 & 77.7 & 67.8 & 66.8 & 77.2 & 26.9 & 58.9 & 90.9 & 92.5 & 69.7 & 42.7 & 48.3 & 96.3 & 49.9 & 66.5 & 39.2 & 29.3 & 17.7 & 50.0 & 68.0 & 47.6 & 19.4 & {53.5} & 53.1 \\ \hline
MoDE{-2} & 61.2 & 68.7 & 84.1 & 95.3 & 78.6 & 69.5 & 67.0 & 80.8 & {30.9} & 60.6 & 91.0 & 92.9 & 71.9 & 40.8 &\better{50.4} & 96.3 & 51.3 & \better{67.9} & 44.2 & \better{31.4} & 18.3 & 51.3 & \better{69.0} & 47.4 & 23.2 & 52.6 & \better{54.4} \\
MoDE{-4} & {61.7} & {68.8} & {85.8} & 95.2 & {79.0} & {74.4} & 67.5 & 83.3 & 29.5 & 60.3 & {91.9} & {92.9} & {72.1} & 49.7 & 46.9 & {96.4} & 50.3 & 66.8 & {51.6} & {28.5} & \better{19.6} & 50.1 & 68.4 & \better{48.3} & 21.6 & 52.6 & 52.2 \\ 
MoDE{-8} & 63.4 & 69.3 & 88.1 & 95.6 & 80.1 & 76.0 & 68.2 & 87.7 & \better{46.7} & \better{60.9} & 91.2 & 93.4 & 77.1 & 46.5 & 47.2 & 97.1 & 58.3 & 67.7 & \better{52.7} & 27.4 & 18.5 & 50.1 & {68.6} & 48.2 & 25.2 & 53.3 & 52.1 \\
MoDE{-16} & \better{64.0} & \better{70.7} & \better{88.4} & \better{96.1} & 80.5 & \better{80.8} & 67.9 & 87.1 & 44.6 & 59.9 & 92.2 & 93.2 & 79.4 & \better{50.1} & {49.8} & 97.1 & \better{60.3} & 67.7 & 48.5 & 26.1 & 18.9 & 55.3 & 68.1 & \better{48.3} & 25.7 & \better{54.1} & 51.9 \\
MoDE{-32} & \better{64.0} & 69.6 & 88.2 & 95.9 & \better{80.8} & 80.1 & \better{68.3} & \better{88.9} & 44.1 & 59.9 & \better{92.6} & \better{93.5} & \better{83.0} & 42.9 & 46.9 & \better{97.4} & 56.2 & 67.3 & 48.8 & {30.7} & {19.2} & \better{58.0} & 68.2 & 48.1 & \better{30.6} & 53.5 & 50.2 \\ 
\shline
\end{tabular}
}
\caption{Performance details of Fig. 3 when scaling MoDE{-n} based on ViT-B/32 on 2.5B image-caption pairs.}
\label{tab:clip2b5-scale}
\end{table*}

\begin{table*}[!htb]
    \centering
    \vspace{-1.em}
    \setlength\tabcolsep{1.0pt}
    \resizebox{\linewidth}{!}
    {\renewcommand{\arraystretch}{1.0}
    \begin{tabular}{c|cccccc|cccccc|cccccc|cccccc}
    \shline
    \multirow{3}{*}{Approach} & \multicolumn{6}{c|}{Text Retrieval}                       & \multicolumn{6}{c|}{Image Retrieval}  & \multicolumn{6}{c|}{Text Retrieval}                       & \multicolumn{6}{c}{Image Retrieval}                      \\
                              &  \multicolumn{3}{c}{COCO} & \multicolumn{3}{c|}{Flickr30k} & \multicolumn{3}{c}{COCO} & \multicolumn{3}{c|}{Flickr30k} & \multicolumn{3}{c}{COCO} & \multicolumn{3}{c|}{Flickr30k} & \multicolumn{3}{c}{COCO} & \multicolumn{3}{c}{Flickr30k}\\ 
                              & R@1      & R@5      & R@10    & R@1    & R@5    & R@10   & R@1      & R@5      & R@10    & R@1    & R@5    & R@10  & R@1      & R@5      & R@10    & R@1    & R@5    & R@10  & R@1      & R@5      & R@10    & R@1    & R@5    & R@10   \\ \hline
    \rowcolor{lightgray} \multicolumn{25}{l}{ViT-B/32}\\
    OpenAI CLIP & 50.2 & 75.0 & 83.5 & \better{78.9} & \better{94.9} & \better{98.2} & 30.4 & 56.0 & 66.9 & 58.8 & 83.6 & 90.0 & -    & -    & -    & -    & -    & -    & -    & -    & -    & -    & -    & -    \\
    OpenCLIP    & 52.5 & 76.8 & 84.7 & 78.8 & 94.1 & 97.0 & 35.3 & 60.9 & 71.7 & 61.7 & 85.5 & 90.9 & 56.3 & 79.8 & 87.1 & \better{84.1} & \better{96.2} & \better{98.3} & 39.3 & 65.4 & 75.6 & 66.7 & 88.4 & 93.1 \\
    \baselinetable{} & 51.8 & 76.4 & 84.7 & 77.8 & 93.5 & 97.1 & 35.9 & 61.8 & 72.1 & 62.3 & 85.5 & 91.5 & 55.2 & 78.9 & 86.5 & 80.7 & 95.2 & 97.3 & 38.1 & 64.1 & 74.3 & 65.1 & 87.7 & 92.7 \\ \hline
    \approach{-2} & 53.3 & 76.7 & 84.8 & 78.6 & 94.3 & 96.9 & 36.4 & 62.1 & 72.6 & 63.0 & 86.1 & \better{91.8} & 56.7 & \better{80.2} & \better{87.5} & 82.8 & 95.1 & 98.2 & 39.5 & 65.3 & 75.3 & 66.4 & 89.0 & 93.6 \\
    \approach{-4} & \better{53.7} & \better{77.2} & \better{85.1} & 78.5 & \better{94.9} & 96.8 & \better{36.7} & \better{62.5} & \better{73.0} & \better{63.6} & \better{86.4} & 91.7 & \better{57.4} & {80.1} & 87.3 & 82.9 & 95.6 & 97.7 & \better{39.9} & \better{66.1} & \better{75.7} & \better{66.7} & \better{88.4} & \better{93.3} \\ \shline
    \rowcolor{lightgray} \multicolumn{25}{l}{ViT-B/16}\\
    OpenAI CLIP & 52.4 & 76.7 & 84.6 & 86.2 & \better{98.0} & \better{99.5} & 33.1 & 58.4 & 69.0 & 69.8 & \better{90.4} & \better{94.6} & -    & -    & -    & -    & -    & -    & -    & -    & -    & -    & -    & -    \\
    OpenCLIP & 55.4 & 79.7 & 86.9 & 83.4 & 96.8 & 98.5 & 38.4 & 63.6 & 73.9 & 65.7 & 88.3 & 93.0 & 59.5 & 81.8 & 88.6 & 86.2 & \better{98.0} & 99.5 & 42.3 & 67.7 & 77.1 & 69.8 & {90.4} & {94.6} \\
    \baselinetable{} & 56.4 & 79.9 & 87.1 & 85.7 & 97.2 & 98.7 & 40.0 & 65.3 & 75.3 & 67.6 & 89.6 & 94.2 & 59.4 & 80.6 & 87.8 & 85.5 & 97.4 & 98.9 & 41.4 & 67.2 & 76.9 & 70.7 & 90.8 & 94.5 \\ \hline
    \approach{-2} & 57.5 & 80.3 & 87.6 & 86.5 & 97.0 & 98.8 & 40.4 & 65.6 & 75.6 & 68.7 & 89.4 & 94.2 & 60.7 & 82.6 & 89.0 & 87.3 & 97.6 & 99.2 & 43.1 & 68.6 & 77.8 & 72.1 & \better{91.8} & 95.3  \\
    \approach{-4} & \better{57.7} & \better{81.1} & \better{88.1} & \better{86.6} & 97.5 & 98.8 & \better{41.0} & \better{66.2} & \better{75.8} & \better{68.7} & 90.0 & 94.2 & \better{62.7} &\better{82.9} & \better{89.8} & \better{89.4} & \better{98.0} & \better{99.6} & \better{44.1} & \better{69.5} & \better{78.7} & \better{72.6} & \better{91.8} & \better{95.4} \\ \shline
    \rowcolor{lightgray} \multicolumn{25}{l}{ViT-L/14}\\
    OpenAI CLIP & 56.3 & 79.4 & 86.6 & 85.2 & 97.4 & 99.2 & 36.5 & 61.0 & 71.1 & 64.9 & 87.2 & 92.0 & -    & -    & -    & -    & -    & -    & -    & -    & -    & -    & -    & -    \\
    OpenCLIP & 59.7 & 82.2 & \better{89.4} & 87.6 & 97.8 & 99.5 & 43.0 & 68.0 & 77.4 & 70.2 & 90.9 & 94.6 & 63.3 & 83.9 & 90.8 & 89.5 & 98.7 & 99.4 & 46.5 & 71.1 & 79.8 & 75.5 & 92.9 & 95.9 \\
    \baselinetable{} & 60.0 & \better{82.9} & \better{89.4} & 86.2 & \better{98.1} & 99.6 & 43.8 & \better{68.7} & \better{77.8} & 73.4 & 92.3 & 95.7 & 64.4 & 85.0 & 91.3 & 90.1 & 98.6 & 99.3 & 47.1 & 71.4 & 80.3 & 76.5 & 93.6 & 96.5 \\ \hline
    \approach{-2} & \better{60.8} & 82.6 & \better{89.4} & \better{87.7} & \better{98.1} & 99.5 & \better{44.2} & 68.6 & \better{77.8} & 73.4 & \better{92.5} & 95.8 & 65.2 & 85.3 & 91.6 & 90.9 & 98.9 & 99.6 & 47.9 & 72.1 & 80.6 & 77.2 & 93.7 & 96.6 \\
    \approach{-4} & 60.6 & \better{82.9} & 89.1 & 86.7 & 97.9 & \better{99.7} & \better{44.2} & 68.6 & \better{77.8} & \better{73.5} & 92.1 & \better{95.9} & \better{65.5} & \better{85.4} & \better{91.8} & \better{91.2} & \better{99.0} & \better{99.7} & \better{48.2} & \better{72.4} & \better{80.7} & \better{77.6} & \better{93.7} & \better{96.7} \\ \shline
    \rowcolor{lightgray} Pre-Train Dataset: & \multicolumn{12}{c|}{400M Image-Caption Pairs} & \multicolumn{12}{c}{OpenCLIP:2.3B, \baselinetable{} / MetaCLIP: 2.5B} \\
    \end{tabular}}
    \caption{\textbf{Zero-shot Retrieval}. The results are separated by the scale of pre-train set. Entries in blue are the best ones.}
    \label{tab:retrieval-full}
\end{table*}

\newpage

\section{Full Results}

Below we provide the complete results of \approach{} reported in \cref{sec:exp} if mentioned. 
Firstly, we compare the performance on CLIP evaluation benchmark, and reports the scores by scaling up the number of coarse-grained clusters in  \cref{tab:clip2b5-scale}. When more data experts are learned, the average accuracy on CLIP benchmark keeps improving.

Secondly, we summarize the results for zero-shot retrieval in \cref{tab:retrieval-full}. 
The results are separated by the scale of pre-train dataset. Consistently, our approach can outperform the \baseline{} in all cases. \approach{} also achieves the best score in most cases.

We noticed the work LiMoE~\cite{mustafa2022multimodal} which follows conventional Deep Mixture of Expert models and trains a stack of Transformer MoE layers on all 3.6B image-caption pairs~\cite{zhai2022lit}. However, the number of parameters in a single LiMoE network is much larger than a single dense baseline. As all of the network parameters are trained synchronously, it will cause huge memory usage. Meanwhile, comparing with \approach{-4} trained on different data clusters while the total pre-train set has only about 2.5B image-caption pairs, our system is more flexible and also achieve better results consistently.

\begin{table}[!htb]
    \centering
    \setlength\tabcolsep{1.0pt}
    \resizebox{0.9\linewidth}{!}{
    \begin{tabular}{c|cccc}
    \shline
                    &        & ~~ViT-B/32~~ & ~~ViT-B/16~~ & ~~ViT-L$^*$~~ \\ \hline
    classification  & LiMoE  & 67.5     & 73.7     & 78.6   \\
       (ImageNet)    & MoDE-4 & 68.9     & 74.3     & 79.4   \\ \shline
    text retrieval  & LiMoE  & 45.7     & 51.3     & 55.7   \\
        (CoCo)     & MoDE-4 & 57.4     & 62.7     & 65.6   \\ \shline
    image retrieval & LiMoE  & 31.0       & 36.2     & 39.6   \\
         (CoCo)       & MoDE-4 & 39.9     & 44.1     & 48.2   \\ \shline
    \multicolumn{5}{l}{\small $^*$: LiMoE uses L/16 and \approach{} uses L/14.}
    \end{tabular}}
    \caption{Comparison with LiMOE~\cite{zhai2022lit}}
    \label{tab:limoe}
\end{table}

\section{Ablation Study Details for Clustering}

Firstly, for ablation details on Clustering Strategy in \cref{sec:cluster-strategy}, we show details in \cref{tab:cluster-supp} for \cref{tab:ensemble-ablate} and \cref{tab:clip400m-scale} for \cref{fig:specturm-fine}.

\begin{table*}[!htb]
\centering
\setlength\tabcolsep{1.0pt}
\resizebox{\linewidth}{!}{
\begin{tabular}{lc|cccccccccccccccccccccccccc}
& \datatag{Average}
& \datatag{ImageNet}
& \datatag{Food-101} 
& \datatag{CIFAR10} 
& \datatag{CIFAR100} 
& \datatag{CUB}
& \datatag{SUN397}
& \datatag{Cars}
& \datatag{Aircraft}
& \datatag{DTD}
& \datatag{Pets}
& \datatag{Caltech-101}
& \datatag{Flowers}
& \datatag{MNIST}
& \datatag{FER-2013}
& \datatag{STL-10}
& \datatag{EuroSAT}
& \datatag{RESISC45}
& \datatag{GTSRB}
& \datatag{KITTI}
& \datatag{Country211}
& \datatag{PCAM}
& \datatag{UCF101}
& \datatag{Kinetics700}
& \datatag{CLEVR}
& \datatag{HatefulMemes}
& \datatag{SST2}\\
\shline
\rowcolor{lightgray} \multicolumn{28}{l}{400M Image-Caption Pairs}\\
\baselinetable{} & 58.2 & 65.5 & 80.6 & \better{91.3} & 70.2 & 63.4 & 63.0 & 70.7 & 26.8 & 52.8 & 88.7 & 91.9 & 68.5 & 41.5 & \better{35.9} & 95.4 & 52.6 & \better{64.2} & 35.8 & 30.7 & 17.2 & 55.5 & 66.1 & 45.4 & \better{30.6} & 56.4 & 53.4 \\
OneStep-2 & 58.0 & 65.0 & 80.4 & \better{91.3} & 69.9 & 62.2 & 62.5 & 69.0 & 27.1 & 52.7 & 88.5 & 91.7 & 67.3 & 40.2 & 32.3 & 95.0 & \better{54.8} & 63.9 & \better{36.2} & \better{36.6} & 16.7 & 54.5 & 66.4 & 45.1 & 26.4 & 57.9 & 54.0 \\
CoarseCluster-2 & 58.5 & 66.1 & 81.1 & 91.0 & 70.6 & 65.3 & 63.1 & 71.8 & 27.1 & 53.5 & 89.0 & 92.2 & 68.7 & \better{45.2} & 33.5 & 95.4 & 52.0 & 63.7 & 34.9 & 34.2 & 17.3 & 54.3 & 66.1 & 45.5 & 29.3 & 56.6 & \better{54.6} \\
\approach{}-2 & 58.6 & 66.1 & 81.2 & 90.9 & 70.5 & 65.2 & 63.0 & 72.0 & 28.3 & 53.5 & 89.4 & \cbetter{92.3} & 68.2 & \better{45.2} & 33.5 & 95.4 & 51.9 & 63.7 & 34.9 & 34.2 & {17.3} & 54.3 & 65.9 & 45.5 & 29.3 & 56.6 & \better{54.6} \\
\hline \hline
CoarseCluster-4 & 58.7 & 66.2 & 82.2 & 91.2 & 70.8 & \better{67.4} & 63.2 & 73.7 & 28.2 & \better{54.0} & 89.7 & 92.2 & 69.8 & 38.1 & 33.2 & 95.6 & 53.5 & 64.0 & 35.2 & 33.8 & \better{17.8} & 53.2 & 66.4 & 45.7 & 29.9 & 57.1 & 53.3 \\
\approach{}-4 & \better{59.0} & \cbetter{66.4} & \better{82.3} & \cbetter{91.3} & \cbetter{70.9} & {67.0} & \better{63.7} & \better{73.8} & \cbetter{30.1} & 52.6 & \cbetter{89.9} & 92.1 & \cbetter{69.2} & 37.9 & 33.2 & \better{95.7} & {53.5} & {64.1} & 35.2 & 33.9 & 17.1 & \better{58.4} & \cbetter{66.6} & \better{45.9} & 30.0 & \cbetter{58.0} & 54.5 \\
\shline
\rowcolor{lightgray} \multicolumn{28}{l}{2.5B Image-Caption Pairs}\\
\baselinetable{} & 59.8 & 67.7 & 82.6 & 95.2 & 77.7 & 67.8 & 66.8 & 77.2 & 26.9 & 58.9 & 90.9 & 92.5 & 69.7 & 42.7 & 48.3 & 96.3 & 49.9 & 66.5 & 39.2 & 29.3 & 17.7 & 50.0 & 68.0 & 47.6 & 19.4 & {53.5} & 53.1 \\ 
OneStep-2 & 59.8 & 67.6 & 82.3 & 94.8 & 77.5 & 67.8 & 66.3 & 76.8 & 26.4 & 58.1 & 90.9 & 92.0 & 68.7 & 45.1 & 47.6 & 96.1 & 50.1 & 65.9 & 43.6 & 29.8 & 17.6 & 49.7 & 67.7 & 47.2 & 20.2 & \better{53.6} & 51.8\\
CoarseCluster-2 & 60.6 & 68.6 & 81.9 & 95.1 & 77.8 & 68.7 & \better{68.0} & 77.8 & 27.3 & 57.2 & 90.3 & 92.6 & 68.4 & 44.7 & 50.3 & 96.3 & 50.7 & 67.2 & 47.1 & \better{33.2} & 18.4 & 50.6 & \better{69.6} & 48.2 & 19.1 & 52.5 & 53.2 \\
\approach{-2} & 61.2 & 68.7 & 84.1 & \better{95.3} & 78.6 & 69.5 & 67.0 & 80.8 & \cbetter{30.9} & \better{60.6} & 91.0 & \better{92.9} & 71.9 & 40.8 &\better{50.4} & 96.3 & \better{51.3} & \cbetter{67.9} & 44.2 & 31.4 & 18.3 & \better{51.3} & {69.0} & 47.4 & \better{23.2} & 52.6 & \better{54.4} \\ \hline \hline
CoarseCluster-4 & 61.3 & \better{69.1} & 83.9 & 95.1 & 78.1 & 73.1 & {67.5} & 82.2 & 27.8 & 60.4 & 90.9 & \better{92.9} & 70.1 & \better{49.7} & 46.9 & 96.2 & 50.4 & 66.4 & 50.4 & 28.4 & 18.8 & 50.0 & 68.8 & 48.1 & 21.6 & 52.7 & 52.9 \\
\approach{-4} & \better{61.7} & {68.8} & \better{85.8} & 95.2 & \better{79.0} & \better{74.4} & {67.5} & \better{83.3} & 29.5 & 60.3 & \better{91.9} & \better{92.9} & \better{72.1} & \better{49.7} & 46.9 & \better{96.4} & 50.3 & 66.8 & \better{51.6} & {28.5} & \better{19.6} & 50.1 & 68.4 & \better{48.3} & 21.6 & 52.6 & 52.2 \\
\shline
\end{tabular}
}
\caption{Performance details for ablation study on clustering strategy in Table 6 (Sec. 5.2). The experiments are performed on ViT-B/32. The results are separated by the scale of pre-train set.}
\label{tab:cluster-supp}
\end{table*}

\begin{table*}[!htb]
\vspace{-1.em}
\centering
\setlength\tabcolsep{1.0pt}
\resizebox{\linewidth}{!}{
\begin{tabular}{lc|cccccccccccccccccccccccccc}
& \datatag{Average}
& \datatag{ImageNet}
& \datatag{Food-101} 
& \datatag{CIFAR10} 
& \datatag{CIFAR100} 
& \datatag{CUB}
& \datatag{SUN397}
& \datatag{Cars}
& \datatag{Aircraft}
& \datatag{DTD}
& \datatag{Pets}
& \datatag{Caltech-101}
& \datatag{Flowers}
& \datatag{MNIST}
& \datatag{FER-2013}
& \datatag{STL-10}
& \datatag{EuroSAT}
& \datatag{RESISC45}
& \datatag{GTSRB}
& \datatag{KITTI}
& \datatag{Country211}
& \datatag{PCAM}
& \datatag{UCF101}
& \datatag{Kinetics700}
& \datatag{CLEVR}
& \datatag{HatefulMemes}
& \datatag{SST2}\\
\shline
\baselinetable{} & 58.2 & 65.5 & 80.6 & 91.3 & 70.2 & 63.4 & 63.0 & 70.7 & 26.8 & 52.8 & 88.7 & 91.9 & 68.5 & 41.5 & \better{35.9} & 95.4 & 52.6 & \better{64.2} & 35.8 & 30.7 & 17.2 & 55.5 & 66.1 & 45.4 & {30.6} & \better{56.4} & 53.4 \\ \hline
$m=2$ & 58.0 & 65.0 & 80.4 & 91.3 & 69.9 & 62.2 & 62.5 & 69.0 & 27.1 & 52.7 & 88.5 & 91.7 & 67.3 & 40.2 & 32.3 & 95.0 & \better{54.8} & 63.9 & 36.2 & \better{36.6} & 16.7 & 54.5 & 66.4 & 45.1 & 26.4 & 57.9 & 54.0 \\
$m=4$ & 58.2 & 65.2 & 81.2 & 91.1 & 70.1 & 62.6 & 63.0 & \better{72.1} & 28.2 & \better{53.5} & 89.1 & 92.1 & 69.0 & 37.7 & 33.5 & 95.3 & 53.0 & 63.6 & 36.0 & 35.5 & \better{17.6} & 54.2 & 65.8 & 45.0 & 28.2 & 56.4 & 55.0 \\  
$m=32$ & 58.4 & 66.0 & 81.2 & \better{91.4} & \better{70.7} & 64.9 & 63.1 & 71.7 & 28.0 & 53.2 & 88.8 & 92.1 & 69.0 & 38.1 & 32.9 & 95.4 & 53.1 & \better{64.2} & 36.3 & 34.0 & 17.4 & 54.1 & 66.0 & \better{45.6} & 29.0 & 56.3 & \better{55.2}\\ 
$m=256$ & 58.4 & 65.9 & 80.6 & 91.2 & 70.2 & 64.2 & 63.5 & 70.5 & 27.2 & 52.6 & 88.8 & 92.2 & 68.5 & 40.0 & 35.2 & 95.3 & {53.5} & 64.0 & \better{39.5} & 34.9 & 17.2 & 53.7 & 66.0 & \better{45.6} & 28.3 & 56.0 & 54.4 \\
$m=512$ & 58.5 & 65.9 & 81.2 & 91.2 & 70.3 & 64.6 & \better{63.6} & 72.0 & 29.0 & 52.5 & 89.2 & 92.0 & 69.7 & 40.0 & {35.3} & \better{95.5} & 52.9 & 63.2 & 39.1 & 34.9 & 17.1 & 53.7 & 66.1 & 45.2 & 27.4 & 56.0 & 54.6\\ 
$m=1024$ & 58.6 & \better{66.1} & 81.2 & 90.9 & 70.5 & 65.2 & 63.0 & 72.0 & 28.3 & \better{53.5} & \better{89.4} & {92.3} & 68.2 & \better{45.2} & 33.5 & 95.4 & 51.9 & 63.7 & 34.9 & 34.2 & {17.3} & 54.3 & 65.9 & 45.5 & 29.3 & {56.6} & 54.6 \\ 
$m=2048$ & \better{58.7} & 65.8 & \better{81.4} & 91.2 & 70.4 & \better{66.1} & 63.3 & \better{72.1} & \better{29.6} & 51.5 & 89.1 & \better{92.4} & \better{70.2} & 43.0 & 33.2 & 95.1 & 53.1 & 63.8 & 32.9 & 32.7 & 17.1 & \better{57.9} & \better{66.7} & 45.2 & \better{31.5} & 56.1 & 54.9 \\ 
\shline
\end{tabular}
}
\caption{Performance details of \approach{-2} when ablating the number of finegrained clusters in Step 1 (Fig. 5 in Sec 5.2). Experiments are performed on ViT-B/32 on 400M image-caption pairs.}
\label{tab:clip400m-scale}
\end{table*}

\begin{table*}[!htb]
\vspace{-1.em}
\centering
\setlength\tabcolsep{1.0pt}
\resizebox{\linewidth}{!}{
\begin{tabular}{lc|cccccccccccccccccccccccccc}
& \datatag{Average}
& \datatag{ImageNet}
& \datatag{Food-101} 
& \datatag{CIFAR10} 
& \datatag{CIFAR100} 
& \datatag{CUB}
& \datatag{SUN397}
& \datatag{Cars}
& \datatag{Aircraft}
& \datatag{DTD}
& \datatag{Pets}
& \datatag{Caltech-101}
& \datatag{Flowers}
& \datatag{MNIST}
& \datatag{FER-2013}
& \datatag{STL-10}
& \datatag{EuroSAT}
& \datatag{RESISC45}
& \datatag{GTSRB}
& \datatag{KITTI}
& \datatag{Country211}
& \datatag{PCAM}
& \datatag{UCF101}
& \datatag{Kinetics700}
& \datatag{CLEVR}
& \datatag{HatefulMemes}
& \datatag{SST2}\\
\shline
\baselinetable{} & 58.2 & 65.5 & 80.6 & 91.3 & 70.2 & 63.4 & 63.0 & 70.7 & 26.8 & 52.8 & 88.7 & 91.9 & 68.5 & 41.5 & 35.9 & \better{95.4} & 52.6 & 64.2 & 35.8 & 30.7 & 17.2 & 55.5 & 66.1 & 45.4 & \better{30.6} & 56.4 & 53.4 \\ \hline
DINOv2 & 58.1 & 65.2 & 80.5 & 91.2 & 70.3 & 63.4 & \better{63.1} & 69.8 & 26.5 & 51.6 & 89.0 & 91.8 & 68.1 & 41.0 & 36.4 & 95.2 & 53.4 & 63.0 & 37.3 & 35.0 & 16.7 & 53.7 & 65.6 & 45.4 & 26.8 & 56.0 & 53.5 \\
Image (CLIP Seed) & 58.3 & 64.7 & 80.6 & 91.3 & \better{70.7} & 63.0 & 63.0 & 70.8 & 27.4 & 53.4 & 87.8 & 92.1 & 68.9 & 42.2 & 33.2 & 95.2 & 53.6 & 62.4 & 38.8 & 34.4 & 16.9 & \better{61.6} & 65.9 & 45.2 & 20.3 & \better{57.8} & 55.6 \\
Image \& Lang. (CLIP Seed) & 58.4 & 65.5 & 80.3 & 91.3 & 70.2 & 63.4 & 63.0 & 70.3 & 27.7 & 52.0 & 88.7 & 91.8 & 68.3 & 40.0 & 35.3 & 95.1 & \better{54.4} & \better{64.4} & 38.9 & \better{36.0} & 16.7 & 54.0 & 66.2 & \better{45.7} & 27.4 & 56.6 & 54.6 \\
Lang. (CLIP Seed) & 58.3 & 65.2 & 80.7 & 91.3 & 69.8 & 64.8 & 62.6 & 71.9 & 26.9 & 52.3 & 88.8 & 91.7 & 68.6 & 39.0 & 34.1 & 95.2 & 54.1 & 63.1 & 38.1 & 33.8 & 16.8 & 54.8 & 66.1 & 45.2 & 27.6 & 57.5 & \better{55.8} \\
SimCSE-UnSup & \better{58.6} & 65.7 & 80.3 & \better{91.4} & 69.6 & 64.4 & {63.0} & 71.8 & 26.6 & 52.0 & 88.9 & 92.1 & \better{69.2} & 41.0 & \better{37.7} & \better{95.4} & \better{54.4} & {64.2} & \better{39.0} & {35.1} & \better{17.3} & 53.5 & \better{66.3} & 45.6 & 26.8 & 56.8 & 55.5 \\
SimCSE-Sup & \better{58.6} & \better{66.1 }& \better{81.2} & 90.9 & {70.5} & \better{65.2} & {63.0} & \better{72.0} & \better{28.3} & \better{53.5} & \better{89.4} & \better{92.3} & 68.2 & \better{45.2} & 33.5 & \better{95.4} & 51.9 & 63.7 & 34.9 & 34.2 & \better{17.3} & 54.3 & 65.9 & 45.5 & {29.3} & 56.6 & 54.6 \\
\shline
\end{tabular}
}
\caption{Performance details on CLIP evaluation benchmark for ablating the embedding types for clustering (Table 7 in Sec. 5.3). The experiments evaluate \approach{-2} based on ViT-B/32 on 400M image-caption pairs.}
\label{tab:ablate-embedding-full}
\end{table*}

Then, for the embedding types, we provide the details of \approach{-2} in \cref{tab:ablate-embedding-full}.
We note that the SimCSE~\cite{gao2021simcse} can be trained via unsupervised or supervised ways. The unsupervised training strategy utilizes dropout masks to generate two views from the same sentence to build positive pair while the latter one uses two sentences which are of similar semantic meaning as positive samples to each other. Regardless the training strategy, we found the average score on CLIP benchmark is the same. 

Meanwhile, when both image and language embeddings are used for clustering, we concatenate their embeddings and we experimentally found that adding the language and image embeddings pair-wisely cannot result in meaningful cluster. However, at inference time, the ensembling weights should be calculated for all image-class pairs in the zero-shot classification task, which is computational heavy but provides very limited gain over the baseline.

\section{Robustness in Training Priority}

For the retrieval-enhanced setup, besides using the class names of a single dataset to retrieve the most important finegrained data clusters, we can also use the class names of all tasks in CLIP benchmark. The detailed results are summarized in \cref{tab:clipeval-prior}.

\begin{table*}[!htb]
\vspace{-1.em}
\centering
\setlength\tabcolsep{1.0pt}
\resizebox{\linewidth}{!}{
\begin{tabular}{lc|cccccccccccccccccccccccccc}
& \datatag{Average}
& \datatag{ImageNet}
& \datatag{Food-101} 
& \datatag{CIFAR10} 
& \datatag{CIFAR100} 
& \datatag{CUB}
& \datatag{SUN397}
& \datatag{Cars}
& \datatag{Aircraft}
& \datatag{DTD}
& \datatag{Pets}
& \datatag{Caltech-101}
& \datatag{Flowers}
& \datatag{MNIST}
& \datatag{FER-2013}
& \datatag{STL-10}
& \datatag{EuroSAT}
& \datatag{RESISC45}
& \datatag{GTSRB}
& \datatag{KITTI}
& \datatag{Country211}
& \datatag{PCAM}
& \datatag{UCF101}
& \datatag{Kinetics700}
& \datatag{CLEVR}
& \datatag{HatefulMemes}
& \datatag{SST2}\\
\shline
\rowcolor{lightgray} \multicolumn{28}{l}{ViT-B/32}\\
OpenAI CLIP     & 56.6 & 63.4 & 83.7 & 89.8 & 65.1 & 53.7 & 62.0 & 59.7 & 19.6 & 44.0 & 87.2 & 87.4 & 66.9 & {48.2} & 46.6 & \better{97.1} & 44.9 & 61.0 & 32.6 & 28.7 & 17.2 & \better{62.5} & 63.9 & {48.0} & 23.6 & \better{56.4} & \better{58.6} \\
OpenCLIP & 61.5 & 66.6 & 82.0 & 93.6 & 75.8 & 66.0 & \better{68.3} & \better{86.0} & 23.9 & 56.1 & 90.5 & 91.9 & 70.5 & \better{70.0} & \better{50.4} & 96.6 & 49.3 & 65.7 & \better{49.3} & {32.7} & 16.7 & {51.7} & 64.9 & 45.6 & \better{24.2} & 52.4 & {57.2} \\
\baselinetable{} & 59.8 & 67.6 & 82.6 & 95.2 & 77.7 & 67.8 & 66.8 & 77.2 & 26.9 & \better{58.9} & 90.9 & 92.5 & 69.7 & 42.7 & 48.3 & 96.3 & 49.9 & 66.5 & 39.2 & 29.3 & 17.7 & 50.0 & 68.0 & 47.6 & 19.4 & {53.5} & 53.1 \\ \hline
Ours & \better{61.9} & \better{70.1} & \better{85.4} & \better{95.7} & \better{80.1} & \better{74.4} & 67.0 & 81.2 & \better{36.4} & 58.5 & \better{91.4} & \better{93.5} & \better{72.7} & 44.7 & 42.2 & {96.8} & \better{53.0} & \better{69.1} & 41.8 & \better{35.8} & \better{18.6} & {58.7} & \better{69.8} & \better{48.9} & 21.7 & 49.7 & 51.3 \\
\rowcolor{lightgray} \multicolumn{28}{l}{ViT-B/16}\\
OpenAI CLIP     & 59.6 & 68.3 & {88.8} & 90.8 & 68.2 & 55.6 & 64.0 & 64.6 & 24.0 & 45.1 & 88.9 & 89.1 & 69.4 & 51.8 & {53.0} & {98.2} & 54.8 & 65.5 & 43.3 & 21.7 & 22.8 & \better{56.3} & 68.5 & {52.3} & 25.5 & \better{58.7} & \better{60.5} \\
OpenCLIP & 62.4 & 70.2 & 86.2 & 94.9 & 76.9 & 70.5 & \better{70.6} & \better{88.2} & 26.6 & 56.3 & 90.4 & 93.1 & 71.0 & 65.8 & {53.3} & 97.9 & \better{55.2} & 68.3 & \better{48.3} & 11.9 & 20.3 & 51.2 & 68.1 & 48.9 & 24.8 & 53.0 & {59.5} \\
\baselinetable{} & 63.5 & 72.1 & 88.3 & 95.7 & 79.0 & 71.4 & 68.5 & 82.9 & 30.3 & \better{62.1} & 91.7 & 93.3 & 73.9 & \better{66.1} & 47.0 & 98.4 & 51.1 & 71.1 & 46.6 & 16.6 & 22.7 & 50.5 & 73.0 & 52.5 & \better{30.8} & 57.4 & 59.0 \\ \hline
Ours & \better{64.8} & \better{74.0} & \better{89.8} & \better{96.3} & \better{81.2} & \better{76.2} & 69.4 & 85.3 & \better{39.1} & 58.4 & \better{92.8} & \better{93.8} & \better{75.9} & 57.4 & \better{48.3 }& \better{98.6} & 54.8 & \better{72.3} & 46.5 & \better{28.0} & \better{23.3} & 50.0 & \better{74.3} & \better{53.4} & 29.2 & 57.8 & 58.4 \\
\rowcolor{lightgray} \multicolumn{28}{l}{ViT-L/14}\\
OpenAI CLIP     & 65.7 & 75.5 & {93.0} & 95.6 & {78.3} & 63.3 & 66.8 & 77.8 & 31.3 & 55.3 & 93.6 & 93.3 & 79.3 & \better{76.4} & \better{56.9} & \better{99.4} & 61.9 & 70.9 & {50.6} & 19.2 &{31.9} & 50.1 & 75.7 & {60.2} & 22.3 & \better{59.7} & \better{68.9} \\
OpenCLIP & 65.7 & 74.0 & 88.6 & 95.8 & 78.3 & 73.5 & 73.5 & \better{91.4} & 34.6 & 61.2 & 92.7 & 93.3 & 74.4 & 64.4 & 53.9 & 98.5 & 58.6 & 71.9 & 51.6 & 26.1 & 24.4 & 58.0 & 73.3 & 52.0 & \better{27.4} & 55.1 & 60.4 \\
\baselinetable{} & 69.8 & 79.2 & 93.4 & 97.6 & 84.2 & 80.1 & \better{73.8} & 88.7 & 44.6 & 68.1 & 94.7 & \better{95.4} & {81.8} & 64.4 & {55.1} & 99.3 & 59.2 & \better{74.6} & 56.3 & \better{29.7} & 34.0 & \better{67.3} & 81.6 & 62.0 & 25.9 & {58.0} & {66.7} \\ \hline
Ours & \better{70.0} & \better{79.4} & \better{93.7} & \better{97.7} & \better{85.0} & \better{81.6} & \better{73.8} & 89.2 & \better{47.5} & \better{68.3} & \better{95.7} & \better{95.4} & \better{83.8} & 69.5 & 52.9 & \better{99.4} & \better{62.4} & 74.1 & \better{59.1} & 29.3 & \better{34.3} & 58.4 & \better{81.8} & \better{62.2} & 23.9 & 57.1 & 65.1 \\
\shline
\end{tabular}
}
\caption{Performance on CLIP evaluation benchmark via in Retrieval-Enhanced setup. The class names of all 26 tasks are jointly used to determine the data clusters. OpenCLIP is trained on LAION-2B with 2.3B image-caption pairs. OpenAI CLIP is trained on WIT400M and its results are included here for complete result summary purpose only.}
\label{tab:clipeval-prior}
\end{table*}

\section{Robustness of Vision Encoders}

For emsembling over model outputs, we can also add the model outputs element-wisely. However, as all vision encoders are separately trained, the learned embedding spaces are not necessarily aligned with each other. As a result, ensembling via element-wise addition does not introduce gain, \eg, for \approach{-4} with ViT-B/16 encoders, the accuracy is only 74.5 compared with 79.6 in \cref{tab:linear_sum}.

Finally, in addition to directly aggregate the feature outputs by all data experts, the parameters learned in \approach{} can also be ensembled via averaging and then used as initialization of a single network for finetuning. As shown in \cref{tab:pav}, we use ViT-B/32 vision encoder, and achieve consistent gain over \baseline{}.

\begin{table}[!htb]
    \centering
    \setlength\tabcolsep{1.0pt}
    \resizebox{0.3\linewidth}{!}{
    \begin{tabular}{c|c}
        \shline
        Approach          & Accuracy \\ \hline
        \baselinetable{} & 73.7     \\
        \approach{-2}    & 74.0       \\
        \approach{-4}    & 74.2     \\
        \approach{-8}    & 73.9     \\
        \approach{-16}   & 74.1     \\
        \approach{-32}   & 74.1     \\
        \shline
    \end{tabular}}
    \caption{Accuracy on ImageNet via parameter averaging.}\label{tab:pav}
\end{table}

\section{Implementation Detail}

\paragraph{Clustering.} We first sample 100M captions from the 400M image-caption pairs to learn the cluster centers in an unsupervised manner. Then, we use nearest neighbor to determine the cluster assignment for all other samples in the 400M as well as 2.5B dataset. We also observed that the cluster centers learned by using less than 2M samples can also result in similar clustering assignments using spherical K-means clustering~\cite{dhillon2001concept} via FAISS~\cite{johnson2019billion}. In practice, we observed that the balanced K-means clustering algorithm does not necessarily enforce strict balance regarding the distribution of the clusters. For example, for the two coarse-grained clusters on 400M dataset used to train \approach{-2}, the number of samples for each cluster are around 170M and 230M respectively. Consequently, as mentioned for Random-2 in \cref{sec:cluster-effect}, mimic the size of subsets by MoDE-2 in the random splitting for fair comparison.

\paragraph{Similarity matrix.} For task-level adaptation, as mentioned in \cref{sec:inference}, we use the nearest neighbor fine-grained cluster ($\argmax_{s\in S} \mathbf{A}_{l, s}$) for each class $l \in L$.  In other words, we apply a maximum filter for each row, \ie, $\mathbf{A}_l$, where the non-maximum values are reset as 0, \ie, $\mathbf{A}_{l, s'}=0$ if $s' \neq \hat{s}$ where $\hat{s} = \argmax_{s\in S} \mathbf{A}_{l, s}$. Then, we scale the raw distance value 5 times, \ie, setting the temperature (divisor) as 0.2, according to our experimental cross validation.

\paragraph{Routing Weights.} As described in \cref{eq:aggregation}, the routing weight $p(c|\mathbf{T})$ of a data expert $f( \cdot | c)$ is essentially obtained via softmax normalization. 
At inference time, we note the routing weights should be reasonably distant from each other.
Consequently, given the classification task with the class names $L$, we use the number of classes $|L|$ to roughly adjust the weights.
Firstly, when $|L|$ is small, \eg, $|L| < 10$, though only one data expert can be activated, the selection could be sensitive to noisy routing. Then, we soften the values in $\mathbf{A}$ by multiplying 
$\exp(0.5-\sqrt{|L|})$ to ensemble two data experts in most cases. 
On the other hand, when $|L|$ is large, \eg, $|L| > 200$, the normalized weights tend to be over-smooth, we thus use a much smaller temperature by dividing the $\lambda$ by $\log(|L|)$. Then, we can only select a few data experts and have low-entropy routing weights.

\end{document}